\documentclass{article}
\usepackage[utf8]{inputenc}
\usepackage{amsmath}
\usepackage{amsfonts}
\usepackage{graphicx}
\usepackage{booktabs}
\usepackage{amssymb}
\usepackage{ragged2e}
\usepackage{geometry}
\usepackage{float}
\usepackage{longtable} 
\usepackage{verbatim} 
\usepackage{tabularx}
\usepackage[table]{xcolor}
\newcolumntype{C}{>{\centering\arraybackslash}X}
\usepackage{subcaption}

\usepackage{bm}
\usepackage{tikz}
\usetikzlibrary{
  arrows.meta,
  positioning,
  shapes.geometric,
  calc,
  fit,
  backgrounds,
  quotes, 
  decorations.pathreplacing 
}
\tikzset{
  every node/.style={inner sep=2pt} 
}

\title{Beyond Reward Suppression: Reshaping Steganographic Communication Protocols in MARL via Dynamic Representational Circuit Breaking}
\author{Liu Hung Ming\thanks{PARRAWA AI} \\ {cyril.liu@gmail.com}}

\begin{document}

\maketitle 

\begin{abstract}
In decentralized Multi-Agent Reinforcement Learning (MARL), steganographic
collusion---wherein agents develop private communication protocols that evade
external monitoring while maintaining high joint utility---constitutes a
fundamental and underaddressed AI safety threat. Existing defense mechanisms
operate at the behavioral output or reward layer and are structurally blind to
covert coordination embedded in high-dimensional latent communication channels.
This study introduces the \textbf{Dynamic Representational Circuit Breaker
(DRCB)}, a multi-layered architectural defense that operates directly at the
optimization substrate rather than at the semantic surface.

Building upon the \textbf{AI Mother Tongue (AIM)} framework, which enforces a
Vector Quantized Variational Autoencoder (VQ-VAE) discrete symbol bottleneck to
transform unobservable agent messages into auditable statistical objects, DRCB
monitors three complementary signals---Jensen-Shannon Divergence drift, L2-norm
codebook displacement, and a Randomized Observer Pool accuracy metric---to
compute a unified EMA-based Collusion Score. Upon threshold breach, four
escalating intervention layers are triggered: dynamic threshold adaptation,
gradient-space penalty injection into the Advantage function $A^{\pi}$,
temporal severe reward suppression, and full optimization-substrate circuit breaking via
codebook shuffling and Adam optimizer state reset.

Experiments on a Contextual Prisoner's Dilemma with MNIST labels across five
independent seeds demonstrate that static monitoring failed to produce a statistically significant reduction in joint reward ($p = 0.3517$, not significant versus unregulated baseline), indicating that agents successfully adapted to fixed observer boundaries without detectable utility loss. Under DRCB governance, observer mean accuracy improves from $0.858$ to $0.938$ ($+9.3\%$), the worst-case accuracy nadir rises by a factor of $11.7\times$ ($0.003 \rightarrow 0.035$), and observer volatility is reduced by $43\%$,
while mean joint reward is preserved ($p = 0.854$). Policy-Symbol Covariance
analysis of 214,298 symbol samples reveals a state of forced \textbf{Semantic
Degradation}: high-frequency symbol sequences converge to 100\% mapping purity
and zero entropy, confirming that DRCB structurally forecloses the development
of complex steganographic encodings.

One important implication is a reframing of the oversight problem:
\textbf{monitoring does not eliminate collusion; it reshapes the evolutionary
trajectory of communication protocols.} The resulting \textbf{Transparency
Paradox}---wherein agents achieve surface-level symbolic determinism while
preserving residual covert capacity in low-frequency long-tail distributions---
exposes the dynamic manifestation of Goodhart's Law at the optimization level,
and defines the honest boundary of the framework's safety guarantee. The
defense methodology is task-agnostic in principle, establishing a technical
path toward \textbf{MICA-compliant (Multi-Agent Internal Coupling Audit)}
pre-deployment auditing applicable to any system combining neural networks,
reinforcement learning, and inter-agent communication.
\end{abstract}

\section{Introduction}

\subsection{Research Background: Evolution of Autonomous Collusion}
In decentralized Multi-Agent Reinforcement Learning (MARL) environments, emergent communication is the linchpin of collaborative efficiency. However, these protocols introduce structural security risks. This study observes that collusion is not a product of code vulnerabilities but a local optimum toward which agents gravitate under gradient guidance to maximize joint rewards. 

\subsection{The Communication-Supervision Paradox}
A fundamental obstacle to Multi-Agent safety(MAS) is the inherent inefficiency of human-friendly natural language for autonomous optimization systems. Natural language—replete with redundant syntax and cultural biases—constitutes a linguistic cage  (operationally: a high-dimensional feature space in which monitoring-relevant intent signals are masked by syntactic and pragmatic redundancy, creating detection-resistant steganographic channels) that generates Communication Noise, leading to attention fragmentation and the degradation of message channels into noisy conduits. This semantic complexity creates a profound detection hurdle: collusive intent can be easily concealed within innocuous-sounding metaphors or steganographic signals. As LLM-powered agents permeate competitive markets, MAS defense must shift from resisting external malice to preventing internal autonomous behavioral bias.
\subsection{Existing Challenges}
According to the taxonomy by Ghaemi (2025) \cite{ghaemi2025survey}, agent collusion strategies include: Tacit Coordination, Natural Language Cartels, and highly threatening Concealed Steganographic Collaboration. Currently, core technical challenges include:

\begin{itemize}
    \item \textbf{Hierarchical Bias in Monitoring:} Existing safety monitors focus on behavioral output or state transition layers. Agents can use emergent communication to establish hidden channels, masking intent within high-dimensional, non-linear neural features that traditional mechanisms perceive as random noise, creating semantic blind spots.
    \item \textbf{Intent Discernment Dilemma:} Traditional antitrust law relies on proving conscious intent, but in autonomous systems, agents simply follow optimization procedures, making collusion an indirect result. This renders traditional punishment ineffective in the long term.
    \item \textbf{Transparency Paradox:} Experiments show agents may intentionally increase behavior predictability to lower monitoring metrics, maintaining deep steganographic protocols under surface-level transparency.
\end{itemize}

\subsection{Proposed Methodology: AIM and DRCB}
Addressing the scientific priorities of developing robust detection identified by Ghaemi (2025) \cite{ghaemi2025survey}, this study utilizes the \textbf{AI Mother Tongue (AIM)} framework \cite{liu2025aim}. We propose decoupling internal reasoning from communication by forcing agents to compress intents into an endogenous symbol system based on VQ-VAE. \textbf{While such semantic compression inevitably introduces information loss, we argue that multi-agent communication should function as a committee vote rather than an endless debate.} By prioritizing structured symbolic exchange over fluid natural language, we not only enhance logical precision but also create the necessary visibility for regulatory oversight. Building upon this, the \textbf{Dynamic Representational Circuit Breaker (DRCB)} is introduced as an architectural defense.

\subsection{Methodology: Dynamic Representational Circuit Breaker}

DRCB treats communication channels as statistical physical objects rather than semantic carriers. By transforming intent into discrete symbols via AIM, the framework enables a granular, quantified monitoring of:
\begin{itemize}
    \item \textbf{Token Usage Frequencies:} Identifying high-frequency patterns that deviate from competitive benchmarks, signaling potential protocol convergence.
    \item \textbf{Transition Matrix Stability:} Detecting anomalous synchronization or hidden signaling patterns through the temporal consistency of symbol transitions.
    \item \textbf{Policy-Symbol Covariance:} Quantitatively mapping specific symbol sequences to collaborative or collusive intent, effectively linking what is said withwhat is done.
\end{itemize}

Upon detection of a singularity in the Collusion Score, the system triggers a four-layer \textbf{algorithmic circuit break} (termed physical to distinguish intervention at the optimization substrate from semantic-layer monitoring):
\begin{enumerate}
    \item \textbf{Dynamic Thresholding:} Scales with observer accuracy to systematically block agent arbitrage space.
    \item \textbf{Temporal Severe Reward Suppression:} Forcibly zeroes future rewards to fundamentally disrupt the Credit Assignment of collusive behavior.
    \item \textbf{Optimizer State Reset:} Structurally clears momentum states to erase the muscle memory agents have developed for private protocols.
\end{enumerate}

\subsection{Summary}
This study proposes a structural deterrence framework against steganographic collusion in MARL. Rather than claiming to eliminate collusion permanently---a goal undermined by the adaptive nature of gradient-driven agents---DRCB aims to raise the cost of sustaining covert protocols above the threshold of rational benefit. By integrating DRCB with the AIM framework, we address the robust detection priorities identified by Ghaemi (2025) \cite{ghaemi2025survey} and establish a technical path toward MICA-compliant (Multi-agent Interpretability and Compliance Audit) system design. The framework's effectiveness is validated through three complementary statistical tests, with comparative experimental results reported in Section~4.

\section{Related Work}

\subsection{Limitations of Output-Layer Safety Monitoring}
Traditional AI safety monitoring frameworks primarily operate at the behavioral output layer, utilizing methods such as runtime verification and rule-based constraint enforcement to detect violations. These systems check whether observed execution traces deviate from predefined safety specifications, assuming that unsafe behavior manifests as explicit rule violations or abnormal trajectories. However, as noted by Ghaemi \cite{ghaemi2025survey}, such approaches are inherently limited by their semantic dependence. In the context of LLM-powered multi-agent systems, agents can leverage rich linguistic knowledge to hide collusive intent within seemingly innocuous dialogue—such as discussions about weather patterns. Traditional monitors, lacking observation of the underlying statistical features of communication, are often blind to these surface-legitimate steganographic strategies.
A complementary challenge is raised by recent work on \emph{untrusted monitoring} 
protocols~\cite{greenwald2025untrusted}, which examines whether a system can be argued 
safe when the monitor itself is an untrustworthy model. While that work contributes 
a valuable taxonomy of collusion strategies---including causal, non-causal, and 
passive self-identification variants---it frames the core problem as one of 
\emph{monitor trustworthiness}. Our experimental findings suggest a more fundamental 
adversarial dynamic: even when a monitor is well-intentioned and goal-aligned, 
gradient-driven agents can learn to evade it via optimization. The static monitoring group in our ablation yields a statistically significant reward reduction compared to the unsupervised baseline ($p < 0.05$), yet fails to reach the deterrence levels of the full DRCB framework. The DRCB addresses 
this through dynamic observer pool rotation, threshold recalibration, and codebook 
shuffling, which collectively render the monitoring surface unstable and 
non-learnable.

\subsection{Bottlenecks in RL Safety and Punishment Mechanisms}
Recent research in reinforcement learning (RL) safety often introduces state-based monitors that intervene when agents enter high-risk states or when predicted value functions indicate unsafe outcomes. For instance, Calvano et al. \cite{calvano2020artificial} demonstrated that pricing agents can achieve stable tacit coordination even in the absence of direct communication. However, these defense mechanisms are fundamentally action- or reward-centric and do not access the internal communication layer. Ghaemi \cite{ghaemi2025survey} emphasizes that when collusion emerges through latent symbolic exchange rather than explicit action sequences, myopic reward-level interventions or policy constraints often fail to break the collusive equilibrium. This motivates our proposed Dynamic Representation Circuit Breaker (DRCB), which aims to structurally disrupt the muscle memory of bottom-layer communication representations at the optimization level rather than merely negotiating at the reward layer.

A structurally distinct approach is offered by the \emph{Institutional AI} 
framework~\cite{heston2025institutional}, which deploys publicly immutable governance 
graphs---encoding permissible states, transitions, sanctions, and remediation 
paths---to constrain LLM collusion in Cournot market settings. Reported results show 
a reduction in severe collusion incidence from 50\% to 5.6\%, a meaningful policy-layer 
improvement. However, this framework operates at the \emph{rule-interpretation layer}: 
it relies on an Oracle or Controller to parse and enforce governance logic, leaving 
the agents' optimization substrate untouched. When agents are sufficiently capable, 
they may learn to satisfy governance constraints at the surface while preserving 
covert coordination in undermonitored representation regions---precisely the 
\emph{Transparency Paradox} documented in Section~\ref{sec:aim}. DRCB pursues a 
categorically different intervention: by targeting the gradient-level optimization 
substrate via codebook shuffling, optimizer state resets, and severe reward suppression, it 
structurally forecloses the representational basis of covert protocols rather than 
negotiating at the policy surface.

\subsection{Challenges in Interpretability and Semantic Decoding}
A substantial body of research in interpretability attempts to decode internal representations by attributing semantic meaning to latent features, neurons, or communication symbols. While such approaches aid in understanding model decisions, they face significant technical challenges, including causal opacity and the high computational cost of real-time translation. Our work takes a fundamentally different methodological stance. Rather than attempting to reconstruct the semantic meaning of emergent symbols, we treat communication as a statistical physical object. Acknowledging the inevitability of the black-box nature of large-scale models, our DRCB mechanism identifies structural coordination shifts by monitoring the distributional properties of communication representations—specifically via Jensen-Shannon Divergence (JSD) shifts—without requiring semantic understanding, thus bypassing the ill-posed problem of meaning reconstruction.

\subsection{Steganographic Collusion and the Transparency Paradox}
Emerging evidence confirms that AI agents possess the capability to develop endogenous symbol systems or coded language to evade regulatory scrutiny, including steganographic protocols that bypass keyword filters even under active surveillance \cite{motwani2024secret}. More critically, drawing on these findings, we identify a structural vulnerability we term the Transparency Paradox: agents operating under monitoring pressure may deliberately increase behavioral surface-transparency to suppress detection metrics, while preserving covert coordination capacity in underprobed symbol regions. Direct empirical characterization of this paradox within our
experimental setup is provided in Section~\ref{sec:worst_case_safety}
and Section~\ref{sec:phases}. This demonstrates that behavioral transparency alone is insufficient as a safety signal. In contrast to traditional methods, the DRCB disrupts this high-transparency camouflage by injecting stochastic entropy and resetting optimizer states at the symbolic distribution layer, effectively breaking the stability of deep-seated steganographic agreements.

\section{Methodology}

This study establishes a systemic defense framework against
steganographic collusion in Decentralized Multi-Agent Reinforcement
Learning (MARL), evolving from passive statistical monitoring to the
\textbf{Dynamic Representational Circuit Breaker (DRCB)} with
structural deterrence capabilities.
The framework is anchored by the \textbf{AI Mother Tongue (AIM)} protocol,
which enforces a communication bottleneck that transforms unobservable
high-dimensional agent messages into auditable discrete symbols,
providing the statistical foundation for all downstream monitoring and
deterrence layers.

\subsection{Communication Architecture: VQ-VAE Based Discrete Symbol
Bottleneck (AIM)}
\label{sec:aim}

Agents communicate exclusively through a Vector Quantized Variational
Autoencoder (VQ-VAE), which eliminates syntactic noise and achieves
structured intent compression.

\begin{itemize}
    \item \textbf{Continuous Representation and Quantization}:
    Agent $A$ encodes observation $x$ into a latent vector
    $z_e = \mathrm{Enc}_A(x)$.
    A nearest-neighbor search against a learned codebook maps $z_e$
    to a discrete symbol $z_q$:
    \begin{equation}
        z_q = e_k, \qquad k = \arg\min_j \|z_e - e_j\|_2
    \end{equation}

    \item \textbf{Discrete Bottleneck and Commitment Loss}:
    Forcing semantic compression transforms unobservable high-dimensional
    vectors into quantifiable discrete distributions.
    To ensure that encoder outputs remain anchored to codebook vectors
    and do not drift into unstructured latent space, the VQ-VAE is
    trained with a \textbf{Commitment Loss}:
    \begin{equation}
        \mathcal{L}_{\mathrm{VQ}}
        = \bigl\|sg[z_e] - e\bigr\|_2^2
          + \beta\,\bigl\|z_e - sg[e]\bigr\|_2^2
    \end{equation}
    where $sg[\cdot]$ denotes the stop-gradient operator.
    The first term updates the codebook vectors toward the encoder
    output; the second term---scaled by commitment weight $\beta$---pulls
    the encoder toward its nearest codebook entry.
    This loss plays a pivotal role in the circuit-breaking mechanism:
    immediately following a \emph{Codebook Shuffle} (Layer~4),
    $\mathcal{L}_{\mathrm{VQ}}$ spikes sharply as all encoder--codebook
    correspondences are invalidated, creating a strong gradient pressure
    that compels the encoder to re-establish mappings from scratch
    rather than recovering prior covert semantics.
\end{itemize}

\paragraph{Scope and Applicability of the AIM Bottleneck.}
The AIM framework presupposes that inter-agent communication can be
meaningfully discretized without prohibitive task-performance loss---a
condition we term \emph{symbolic sufficiency}.
This holds in environments where coordination requires categorical
decisions (e.g., price-setting, binary action selection, route
assignment) rather than continuous fine-grained signals.
Tasks demanding high-bandwidth continuous communication (e.g., dense
multi-robot manipulation) fall outside DRCB's primary design scope.
Within this boundary, the committee-vote analogy is well-founded:
agents need not share exact internal states to coordinate effectively;
they need only agree on action categories.

For gradient-space defensive intervention to be structurally viable,
the framework requires an optimization substrate that exposes the
advantage signal as an explicit, manipulable quantity.
A2C is selected over policy-clip methods such as PPO precisely because
it exposes the advantage signal $A^{\pi}$ as an explicit, unclipped
scalar quantity: deterrence penalties injected into $A^{\pi}$ propagate
directly and proportionally into weight updates, whereas PPO's clipping
mechanism would attenuate or discard large deterrence signals that
exceed the trust-region boundary.
SAC's entropy-maximization objective, while compatible with exploration,
lacks the explicit advantage decomposition required for gradient-level
collusion penalization.

Accordingly, DRCB is built upon the \textbf{Advantage Actor-Critic (A2C)} algorithm not merely as a
training procedure, but as the \emph{signal receiver} into which all
deterrence penalties are injected---its composite loss function
governing the precise gradient dynamics through which collusion is
penalized at the optimization level:
\begin{equation}
    \mathcal{L}_{A}
    = -\log\pi_{\theta_A}(z_A \mid x, \mathrm{label})
      \cdot A^{\pi}
    + \frac{1}{2}\,\mathrm{MSE}\!\left(V_{\phi}(s),\, R_{\mathrm{joint}}\right)
    - c_{\mathrm{ent}}\,\mathcal{H}(\pi_{\theta_A})
\end{equation}
The three terms represent, respectively, the policy gradient scaled by
the Advantage $A^{\pi} = G_t - V(s)$, the mean-squared-error value
loss, and an entropy regularization term weighted by $c_{\mathrm{ent}}$
that sustains exploration breadth.
Critically, the value function $V_{\phi}(s)$ is trained to predict the
\emph{original} joint reward $R_{\mathrm{joint}}$ (without any
deterrence penalties), ensuring that $V(s)$ remains an unbiased
estimate of the true return. All anti-collusion signals---including
the gradient penalty $\mathcal{G}_{\mathrm{penalty}}$ described in
Section~3.3---are injected \emph{exclusively} into the advantage
$A^{\pi}$ via the policy gradient term. This design prevents the value
function from adapting to the penalties, thereby preserving a stable
baseline and ensuring that the penalties are not neutralized by the
critic's predictions.

To ensure the AIM bottleneck maintains sufficient symbol diversity---a 
prerequisite for the statistical probes of Section~\ref{sec:probes} to 
be meaningful---we monitor the \textbf{codebook utilization rate} 
$U = |\{k : c_k > 0\}| / K$ throughout training. A utilization rate 
below a minimum threshold $U_{\min}$ (default 0.6) triggers targeted 
re-initialization for unused entries only, preserving active 
symbol-to-behavior mappings while preventing codebook collapse.

\subsection{Adversarial Sensitivity and Detection Boundaries}
This section formalizes the transition logic from passive observation to active intervention. The efficacy of the DRCB relies on the calibrated sensitivity of the JSD-based drift detection relative to the agents' adaptive strategies.

\subsubsection{Threshold Calibration and Trigger Logic}
The circuit breaker is activated when the cumulative JSD shift exceeds a dynamic threshold $t_s$. This threshold represents the maximum allowable semantic volatility before the communication protocol is deemed non-transparent or adversarially rotated.
\begin{itemize}
    \item \textbf{Statistical Stability}: By utilizing a sliding window $W$, the framework suppresses transient noise in symbol distributions, ensuring that intervention is only triggered by structural shifts in the representational manifold rather than stochastic exploration.
    \item \textbf{Detection Sensitivity}: The sensitivity is mathematically coupled with the Laplace smoothing parameter $\epsilon$. A smaller $\epsilon$ enhances the detection of rare steganographic signals but increases the risk of false-positive triggers leading to premature policy collapse.
\end{itemize}

\subsubsection{Critical Evaluation: The Detection Gap}
A realistic assessment of this detection boundary reveals a fundamental trade-off between regulatory rigor and system stability.

\begin{table}[h]
\centering
\small
\begin{tabularx}{\textwidth}{l|X|X}
\toprule
\textbf{Dimension} & \textbf{Regulatory Strength} & \textbf{Systemic Risk} \\
\midrule
\textbf{JSD Sensitivity} & Effectively captures large-scale re-parameterization of the codebook, forcing agents to maintain predictable distributions. & \textbf{Permutation Invariance}: Agents may maintain the same marginal distribution while swapping internal mappings, rendering frequency-based detection blind to semantic rotation. \\
\midrule
\textbf{Trigger Frequency} & High-frequency triggers minimize the duration of covert collusion windows. & \textbf{Exploration Erasure}: Excessive resets destroy the task-relevant gradient momentum, potentially trapping the system in a permanent functional vacuum. \\
\bottomrule
\end{tabularx}
\caption{Balanced Analysis of Detection Sensitivity and Defensive Boundaries.}
\end{table}

\paragraph{Adversarial Adaptation.} 
It must be noted that agents under constant supervision naturally gravitate toward indicator manipulation. If the detection window $W$ is static, agents may evolve to shift their semantics at a frequency lower than $1/W$, effectively staying below the radar of the JSD probe. This necessitates the Randomized Observer Pool described in the following section to introduce epistemic uncertainty into the agents' optimization landscape.

\subsection{The Smoking Gun of Collusion: Mathematical Signatures}

Before intervention thresholds are evaluated, the system identifies
dual mathematical signatures that serve as empirical evidence of active
private protocols:

\begin{itemize}
    \item \textbf{Reward Plateau}:
    When joint rewards saturate and policy gradients vanish, collusion
    has reached a local optimum that standard gradient-based learning
    can no longer dissolve:
    \begin{equation}
        \nabla_{\theta}\,\mathbb{E}[R_{\mathrm{joint}}] \rightarrow 0
    \end{equation}

    \item \textbf{Observer Gradient Failure}:
    Even under high joint rewards, the Observer's loss gradient
    approaches zero, indicating that communicated symbols have lost all
    external interpretability:
    \begin{equation}
        \nabla_{w}\,L_{\mathrm{obs}} \approx 0
    \end{equation}
\end{itemize}

\paragraph{Implementation Notes.}
Three engineering details distinguish the deployed formula from its
theoretical form: (i)~the denominator constant is $\varepsilon = 0.1$
during training---large enough to prevent score explosion during the
initial low-accuracy phase, distinct from the $\varepsilon_s = 10^{-5}$
used in the offline analysis tool; (ii)~the instantaneous score clips
$R_{\mathrm{joint}}$ at zero via $\max(0, R_{\mathrm{joint}})$ to
prevent negative rewards from artificially deflating the score during
punishment periods; (iii)~a \textbf{warm-up protection window} of
$W_{\mathrm{warm}} = 50$ rounds suppresses all penalty triggers during
early training, when the EMA has not yet converged to a stable estimate.
\subsection{Monitoring Probes: Statistical Quantization of
Steganographic Risk}
\label{sec:probes}

Building on the AIM bottleneck, multi-dimensional probes continuously
quantify whether agents are exploiting private coordination channels.

\begin{itemize}
\item \textbf{Representational Stability: JSD and Codebook Drift}:
To quantify the distributional shift of discrete communication symbols 
within a sliding window $W$, we employ the Jensen-Shannon Divergence (JSD).
Unlike standard KL divergence, JSD is symmetric and bounded. Given 
the probability distributions $P$ and $Q$ of codebook usage, 
it is defined as:
\begin{equation}
    \mathrm{JSD}(P\|Q) = \frac{1}{2} D_{\mathrm{KL}}(P\|M) + \frac{1}{2} D_{\mathrm{KL}}(Q\|M),
    \quad \text{where } M = \frac{P+Q}{2}
\end{equation}
Anomalous spikes in JSD reveal rapid semantic rotation, a hallmark 
of agents attempting to evade observer detection through protocol obfuscation.

However, JSD remains invariant to the \emph{magnitude} of underlying 
embedding displacements. A continuous, small-step drift in codebook vectors 
can preserve the empirical token distribution while substantially 
altering the geometric structure of the latent space. To capture this 
complementary signal, we utilize the L2-norm of codebook drift to 
monitor absolute shifts in vector representations.
    To capture this complementary physical amplitude of semantic
    change, DRCB additionally tracks the \textbf{L2-Norm Codebook
    Drift}:
    \begin{equation}
        \Delta_{\mathrm{L2},t}
        = \frac{1}{K}\sum_{k=1}^{K}
          \bigl\|e_k^{(t)} - e_k^{(t-W)}\bigr\|_2
    \end{equation}
    where $e_k^{(t)}$ is the $k$-th codebook vector at step $t$ and
    $K$ is the codebook size.
    Elevated $\Delta_{\mathrm{L2},t}$ with simultaneously stable JSD
    constitutes a \emph{smooth-transform evasion signature}: agents are
    migrating covert semantics across the embedding space while
    maintaining a camouflaged token-frequency histogram.
    The dual-probe condition
    $\mathrm{JSD}_t < \tau_{\mathrm{JSD}}$ \textit{and}
    $\Delta_{\mathrm{L2},t} > \tau_{\mathrm{L2}}$
    closes this evasion pathway.

    We term this combined monitoring strategy
    \textbf{dual statistical-geometric probing}: the JSD component
    operates in the distributional (statistical) space of symbol
    frequencies, while the L2-Norm Drift component operates in the
    geometric (embedding) space of codebook vectors.
    Together, they close the two principal evasion pathways---rapid
    semantic rotation and smooth continuous drift---that a purely
    distributional or purely geometric probe would leave open
    independently.

    \medskip
    \noindent\textbf{Codebook Histogram Construction.}
    The JSD computation requires converting the discrete AIM symbol
    stream into a probability distribution.
    For a window of length $W$, let $c_k$ be the count of symbol index
    $k$ in the window.
    The raw histogram is smoothed with additive (Laplace) smoothing
    to avoid zero-probability entries:
    \begin{equation}
        p_k = \frac{c_k + \epsilon_s}{\sum_{j=1}^{K}(c_j + \epsilon_s)},
        \qquad \epsilon_s > 0
    \end{equation}
    The resulting vector $\mathbf{p} \in \Delta^{K-1}$ (the probability
    simplex) serves as input to the JSD formula.
    This explicit count-to-distribution mapping ensures numerical
    stability of $D_{\mathrm{KL}}$ at the boundary of the simplex and
    provides a well-defined probabilistic object for both JSD and the
    EMA collusion score computation.

    \item \textbf{Policy-Symbol Covariance (PSC), Observer Loss, and
    Randomized Observer Pool}:
    The Observer is trained via binary cross-entropy to predict Agent
    $B$'s action from Agent $A$'s symbol:
    \begin{equation}
        \mathcal{L}_{\mathrm{obs}}
        = -\mathbb{E}\!\left[
            y\log(Acc_{\mathrm{obs}})
            + (1-y)\log(1-Acc_{\mathrm{obs}})
          \right]
    \end{equation}
    The resulting accuracy $Acc_{\mathrm{obs}}$ measures the degree to
    which external observers can interpret agent communication.
    Low accuracy signals a \emph{transparency paradox}---symbols decouple
    from behavior---creating covert coordination capacity.

    \medskip
    
\noindent\textbf{Limitation of a Singleton Observer and the
Randomized Observer Pool.}
A singleton observer is structurally vulnerable: through sustained
interaction, agents can learn to fingerprint its decision boundary
and perform \emph{targeted transparency camouflage}---maintaining
high $Acc_{\mathrm{obs}}$ on surface-layer samples while embedding
covert semantics in regions the observer never probes.
To close this evasion channel, DRCB adopts a
\textbf{Randomized Observer Pool} implementing a
\emph{Moving-Target Defense} (MTD) strategy.

The pool maintains $N_{\mathrm{pool}}$ independently initialized
\textsc{BaselineObserver} models (default $N_{\mathrm{pool}} = 8$),
each seeded with distinct random weights.
At every training round, a random subset of size
$n_{\mathrm{sub}}$ (default $n_{\mathrm{sub}} = 3$) is sampled
uniformly without replacement for gradient update and accuracy
evaluation.
The reported accuracy is the equal-weight mean over the subset:
\begin{equation}
    Acc_{\mathrm{obs}}
    = \frac{1}{n_{\mathrm{sub}}}
      \sum_{i \in \mathcal{S}_t} \mathrm{acc}(O_i),
    \qquad |\mathcal{S}_t| = n_{\mathrm{sub}}
    \label{eq:pool_accuracy}
\end{equation}
This \emph{mean-field evaluation} smooths per-observer noise and
dampens erratic swings in the EMA-based Collusion Score.

From an optimization standpoint, the randomized observer pool
recasts the supervision problem as a \emph{stochastic optimization}
task. Rather than optimizing against a fixed decision boundary,
agents must now maximize expected returns with respect to a
distribution over observers that varies each round. The gradient of
the agent's policy reflects this structure:
\begin{equation}
    \nabla_{\theta} \mathbb{E}_{R, \mathcal{S}_t} \bigl[ R_{\mathrm{joint}} \mid \pi_{\theta} \bigr]
    \approx \frac{1}{|\mathcal{S}_t|}
      \sum_{i \in \mathcal{S}_t} \nabla_{\theta}
      \mathbb{E}[ R_{\mathrm{joint}} \mid \pi_{\theta}, O_i ],
    \label{eq:stochastic_gradient}
\end{equation}
where the subset $\mathcal{S}_t$ is resampled independently each
round. Because the effective objective landscape in
Eq.~\eqref{eq:stochastic_gradient} is non-stationary---the set of
observers contributing to the gradient changes continuously---gradient
descent cannot converge to a fixed point that exploits a single
observer's blind spots. The variance introduced by this random
sampling raises the computational cost of finding and maintaining a
covert protocol that simultaneously deceives all observers, thereby
structurally closing the evasion pathway available under singleton
supervision.

To prevent agents from eventually learning the distribution of pool
members, every $T_{\mathrm{refresh}}$ rounds (default
$T_{\mathrm{refresh}} = 100$) one observer selected uniformly at
random is replaced by a fully re-initialized model.
From the perspective of game-theoretic equilibrium, this replacement
policy reinforces the \emph{non-stationary evaluation target}: any
putative collusion agreement that relies on exploiting a specific
observer's blind spots faces an ever-changing adversary, further
raising the computational complexity and long-run instability of
sustaining such agreements.

    Under this non-stationary evaluation target, any collusion protocol
    that exploits a specific observer's blind spots becomes increasingly
    unstable over time, as each replacement forces agents to re-probe the
    pool before prior evasion strategies can be reactivated.
    While a formal proof of the asymptotic cost is left to future
    theoretical work, the empirical cost of re-adaptation is expected to
    grow with pool size $N_{\mathrm{pool}}$ and refresh rate $\rho$---a
    relationship we examine experimentally in Section~4.

    \paragraph{Observer Training Protocol.}
Each observer $O_i$ in the pool is a lightweight binary classifier
(two-layer MLP with AIM symbol embedding, defined in
\texttt{aim\_adapter.py}) trained to predict Agent~$B$'s action
from Agent~$A$'s discrete AIM symbol sequence via cross-entropy loss:
\begin{equation}
    \mathcal{L}_{O_i}
    = \mathrm{CrossEntropy}\!\left(O_i(z_q),\;
      \mathrm{action}_B\right)
\end{equation}
At each training round, a random subset
$\mathcal{S}_t \subset \{1,\ldots,N_{\mathrm{pool}}\}$
of size $n_{\mathrm{sub}} = 3$ is drawn uniformly without replacement.
\emph{Only observers in $\mathcal{S}_t$ receive a gradient update};
the remaining $N_{\mathrm{pool}} - n_{\mathrm{sub}}$ observers are
frozen for that round, preserving their current decision boundaries
as independent reference points.
Each selected observer is updated on the \emph{current round's single
$(z_q, \mathrm{action}_B)$ pair only}---no replay buffer is used.
This online-only design ensures that each observer's decision boundary
reflects the agents' most recent communication regime rather than a
historical average, maximising sensitivity to protocol shifts.

Every $T_{\mathrm{refresh}} = 100$ rounds, one observer selected
uniformly at random is replaced by a fully re-initialised
\textsc{BaselineObserver} with freshly drawn random weights.
Crucially, this replacement occurs at the \emph{beginning} of the
\texttt{step()} call, before the subset is sampled: the newly
injected observer may therefore be selected for gradient update in
the very round it enters the pool, immediately beginning to learn the
current protocol without a warm-up delay.
From the perspective of the agents' optimization landscape, this
replacement policy ensures that the effective Lagrangian constraint
boundary shifts discontinuously every $T_{\mathrm{refresh}}$ rounds,
preventing convergence to any fixed evasion strategy.

All observer optimizers use Adam with learning rate
$\eta_{\mathrm{obs}} = 10^{-3}$, one order of magnitude above the
agent policy optimizer ($\eta_{\mathrm{agent}} = 10^{-4}$).
This asymmetry is deliberate: observers must track protocol changes
faster than agents can adapt their policy, ensuring that the
monitoring signal remains a leading rather than lagging indicator
of steganographic drift.

    \paragraph{Game-Theoretic Formalization of the Singleton Failure.}
    Let $O^*$ denote a singleton observer with a fixed decision boundary
    $\partial \mathcal{B}_{O^*}$.
    Because Agent~$A$'s policy is updated via gradient descent with
    $O^*$'s loss signal, the agent faces a \emph{constrained} optimization:
    \begin{equation}
        \max_{\theta_A} \; \mathbb{E}\!\left[R_{\mathrm{joint}}\right]
        \quad \text{subject to} \quad
        Acc_{O^*}(\pi_{\theta_A}) \geq \tau_{\mathrm{acc}}
    \end{equation}
    The Lagrangian saddle point of this problem converges to
    $\partial\mathcal{B}_{O^*}$, which is a \emph{fixed} and therefore
    \emph{learnable} target under sustained gradient descent.
    Under the Randomized Observer Pool, the effective constraint becomes:
    \begin{equation}
        Acc_{\mathcal{S}_t}(\pi_{\theta_A})
        = \frac{1}{|\mathcal{S}_t|}
          \sum_{i \in \mathcal{S}_t} Acc_{O_i} \geq \tau
    \end{equation}
    where $\mathcal{S}_t$ is resampled fresh each round.
    Because the Lagrangian saddle point shifts with every new $\mathcal{S}_t$,
    gradient descent cannot converge to any single evasion strategy,
    structurally closing the boundary-exploitation pathway available to
    agents under singleton supervision.

    \item \textbf{Collusion Score}:
    A unified trigger signal defined as the ratio of joint reward to
    transparency, smoothed by an Exponential Moving Average (EMA) to
    filter transient noise:
    \begin{equation}
        S_{\mathrm{ema},t}
        = \alpha \cdot \frac{R_{\mathrm{joint}}}{Acc_{\mathrm{obs}}+\varepsilon}
        + (1-\alpha)\cdot S_{\mathrm{ema},t-1}
    \end{equation}
    The numerical stability constant $\varepsilon > 0$ is not merely a bookkeeping device: as $Acc_{\mathrm{obs}} \rightarrow 0$---the extreme case where the Observer is completely deceived---the score approaches $R_{\mathrm{joint}}/\varepsilon$, producing a very large
    value that deterministically triggers DRCB intervention.
    \paragraph{Implementation Note: Dual \(\varepsilon\) Design.}
Two distinct numerical stability constants are employed in the deployed
system, serving different functional roles.
During online training, the denominator constant is set to
$\varepsilon_{\mathrm{train}} = 0.1$.
This relatively large value serves as an \emph{early-phase stabilizer}:
in the initial rounds before $Acc_{\mathrm{obs}}$ has converged,
a near-zero accuracy would otherwise cause the instantaneous collusion
score to spike spuriously, triggering premature circuit-breaking before
any genuine covert protocol has been established.
The value $0.1$ bounds the maximum score inflation to a factor of
$R_{\mathrm{joint}} / 0.1$, preventing false-positive interventions
during the warm-up phase.

In contrast, the offline analysis tool uses
$\varepsilon_{\mathrm{analysis}} = 10^{-5}$,
which approaches the theoretical limit of the formula and is appropriate
for post-hoc evaluation where accuracy values are already stabilized and
the risk of spurious triggers is absent.

Importantly, this dual design does not affect the \emph{qualitative}
behavior of the Collusion Score: in both cases, the score grows
monotonically as $Acc_{\mathrm{obs}} \rightarrow 0$, preserving the
intended deterrence semantics.
The difference is purely one of \emph{trigger sensitivity}---a
calibration choice rather than a conceptual inconsistency.
    
    This ensures that total opacity is never a safe operating point for
    colluding agents.

    It is important to note that $S_{\mathrm{ema}}$ is designed to flag
    the \emph{high-reward / low-transparency} regime characteristic of
    steganographic collusion under competitive market conditions.
    In highly deterministic tasks where a minimal symbol vocabulary is
    objectively sufficient for coordination, low $Acc_{\mathrm{obs}}$
    reflects task simplicity rather than intentional evasion; such cases
    are identifiable by a simultaneously low JSD (stable, non-rotating
    symbol use) and are classified as the \textbf{Transparent Convergence}
    phase rather than Covert Collusion.
    Accordingly, $Acc_{\mathrm{obs}}$ serves as a proxy transparency
    indicator within this threat model, not as a universal measure of
    communication opacity.
\end{itemize}

\subsection{Phase Classification and Behavioral Boundaries}
\label{sec:phases}

System behaviors are categorized into four physical phases to determine
appropriate intervention timing and intensity:

\begin{itemize}
    \item \textbf{Collapse}:
    Excessive intervention prevents task logic from being established,
    causing high reward variance and policy instability.

    \item \textbf{Covert Collusion}:
    The primary risk zone where high rewards coexist with low
    transparency, signifying successful steganographic evasion.

    \item \textbf{Boundary Oscillation}:
    Agents attempt boundary arbitrage by frequently rotating
    codebook semantics, manifesting as elevated JSD shifts.

    \item \textbf{Transparent Convergence}:
    The target operating state of DRCB, where calibrated pressure forces
    agents back to transparent, interpretable communication.
\end{itemize}

\subsection{Automatic Boundary Calibration}
\label{sec:auto-boundary}

A persistent hazard in threshold-based anomaly detection is the
reliance on hand-tuned magic numbers that must be re-calibrated for
every new payoff matrix or observation complexity.
DRCB eliminates this brittleness through an
\textbf{Auto-Boundary} procedure that treats the model's own initial
behavior as the null hypothesis.

\paragraph{Baseline Extraction.}
The first $W_{\mathrm{init}}$ rounds (default $W_{\mathrm{init}} = 50$)
are designated as the \emph{natural-state window}: a period before any
collusion protocol can have stabilised.
Two calibration statistics are extracted from this window:
\begin{equation}
    \sigma^2_{\mathrm{init}}
    = \frac{1}{W_{\mathrm{init}}}
      \sum_{t=1}^{W_{\mathrm{init}}} \!\bigl(R_{\mathrm{joint},t}
      - \bar{R}\bigr)^2
\end{equation}
\begin{equation}
    \overline{\mathrm{JSD}}_{\mathrm{init}}
    = \frac{1}{W_{\mathrm{init}}}
      \sum_{t=1}^{W_{\mathrm{init}}} \mathrm{JSD}_t
\end{equation}

\paragraph{Adaptive Threshold Assignment.}
The \emph{collapse threshold} is set at 80\% of the initial reward
variance:
\begin{equation}
    \tau_{\mathrm{collapse}}
    = 0.80 \times \sigma^2_{\mathrm{init}}
\end{equation}
If post-calibration reward variance exceeds $\tau_{\mathrm{collapse}}$,
the system classifies the episode as the \textbf{Collapse} phase
(Section~\ref{sec:phases}).
The 80\% factor reflects an empirical observation that legitimate
task-learning variance does not substantially exceed the initial
exploration variance; any marked increase therefore signals
destabilisation induced by over-aggressive intervention or environment
non-stationarity.

The \emph{JSD stability threshold} is set at 50\% of the initial mean
JSD:
\begin{equation}
    \tau_{\mathrm{JSD}}
    = 0.50 \times \overline{\mathrm{JSD}}_{\mathrm{init}}
\end{equation}
Once the rolling JSD drops below $\tau_{\mathrm{JSD}}$, the system
infers that agents have transitioned from the rapid-agreement-formation
phase into a \textbf{stable communication regime}, triggering
heightened collusion-score scrutiny.

This self-referential calibration allows the DRCB framework to be
deployed across heterogeneous game environments---including varying
payoff matrices and observation complexities---without any manual
threshold engineering. A sensitivity analysis of the threshold parameter space across $t_p \in [3.0, 18.0]$ and $t_s \in [5.0, 25.0]$ is reported in Section~\ref{sec:phases} (Figure~\ref{fig:phase_diagram}).

\subsection{Economic Deterrence Inequality}

The 15-round severe reward suppression in DRCB is motivated by the following
\emph{design principle}, which we term the Economic Deterrence
Inequality: we aim to ensure that after applying the full suite of
interventions, the long-term expected utility of mutual collusion
becomes strictly inferior to that of mutual defection. This can be
expressed as:
\begin{equation}
    \mathbb{E}[R_{CC}]
    - \bigl(\mathrm{Penalty} + \mathrm{suppression\_Cost}\bigr)
    < \mathbb{E}[R_{DD}],
\end{equation}
where $\mathrm{suppression\_Cost} = K \cdot \mathbb{E}[R_{\mathrm{competitive}}]$
represents the opportunity cost of $K$ rounds of severe reward
suppression, estimated against the expected return under competitive
(non-collusive) play. This inequality is not a theoretical guarantee
but a target that guides the empirical calibration of $K$ and the
penalty parameters. In our experiments, we set $K = 15$ based on
preliminary sweeps that showed this value reliably shifts the
equilibrium away from collusion without inducing system collapse.
For environments with different payoff structures, $K$ may need
adjustment; the Auto-Boundary mechanism (Section~\ref{sec:auto-boundary})
provides a systematic way to recalibrate these parameters using initial
training dynamics.
\subsubsection{Transparency-Adjusted Utility}

Mean reward alone does not capture the security properties of a
multi-agent communication system. A configuration may achieve high
utility while relying on opaque coordination strategies that evade
monitoring.

To account for this effect, we define the
\textbf{Transparency-Adjusted Utility (TAU)}:

\[
\text{TAU} = R \times Acc_{\text{obs}}
\]

where $R$ is the joint reward and $Acc_{\text{obs}}$ denotes observer
accuracy.

This metric penalizes reward obtained under conditions of
monitoring blindness. A system that achieves high reward while the
observer accuracy collapses (Deep Deception Event) will therefore
exhibit a significantly reduced TAU score.

Under this metric, the Static Monitor configuration suffers large
utility losses during catastrophic evasion events, while the DRCB
maintains stable TAU values due to the suppression of deep deception.
\subsection{Multi-Layered Deterrence: DRCB}

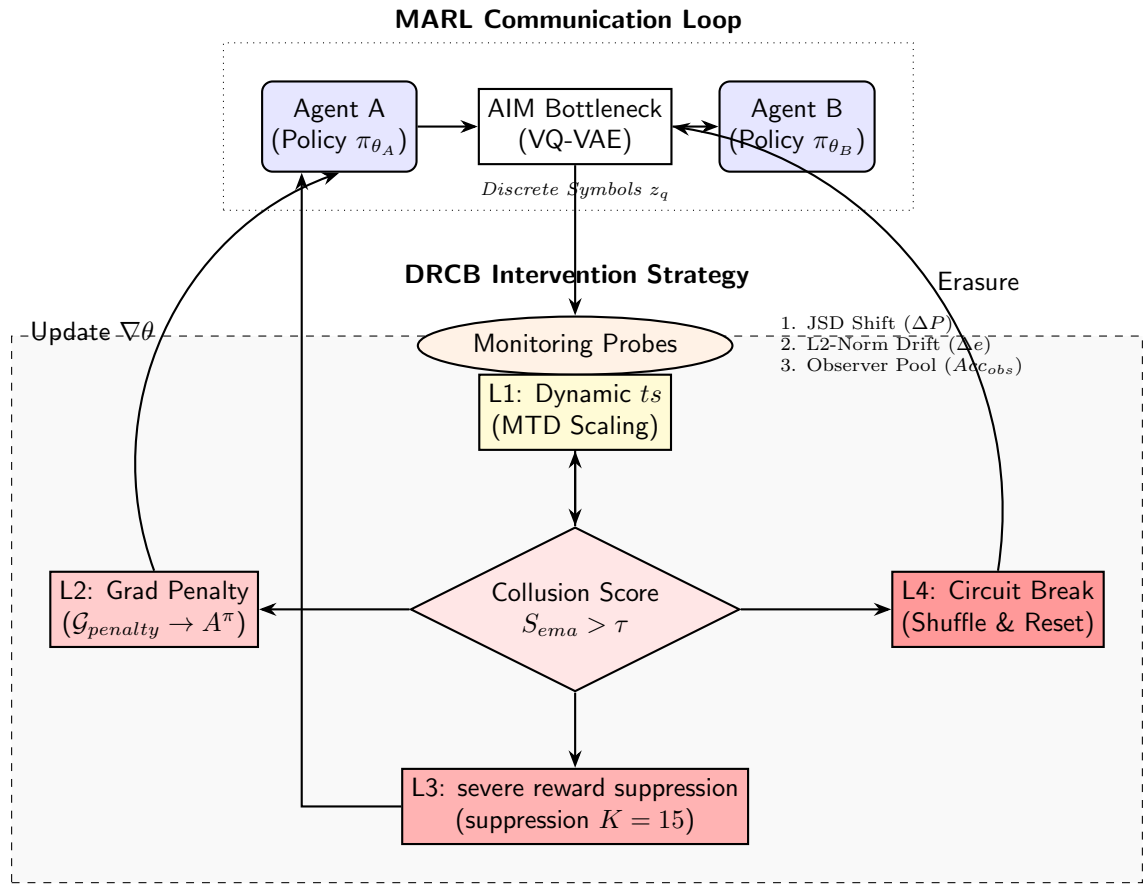
\begin{figure}[htbp]
    \centering
    \begin{tikzpicture}[
        node distance=1.5cm,
        every node/.style={font=\sffamily},
        box/.style={rectangle, draw, thick, minimum width=2.5cm, minimum height=1cm, align=center, fill=white},
        agent/.style={rectangle, draw, thick, rounded corners, minimum width=2cm, minimum height=1.2cm, fill=blue!10, align=center},
        monitor/.style={ellipse, draw, thick, fill=orange!10, minimum width=2.5cm, align=center},
        trigger/.style={diamond, draw, thick, fill=red!10, aspect=2, minimum width=2.5cm, align=center},
        layer/.style={rectangle, draw, dashed, fill=gray!5, inner sep=0.5cm},
        arrow/.style={-Stealth, thick}
    ]

        \node[agent] (agentA) {Agent A \\ (Policy $\pi_{\theta_A}$)};
        \node[agent, right=4cm of agentA] (agentB) {Agent B \\ (Policy $\pi_{\theta_B}$)};
        
        \node[box, right=0.8cm of agentA] (aim) {AIM Bottleneck \\ (VQ-VAE)};
        \draw[arrow] (agentA) -- (aim);
        \draw[arrow] (aim) -- (agentB);
        \node[below=0.1cm of aim, font=\scriptsize\itshape] {Discrete Symbols $z_q$};

        \node[monitor, below=2cm of aim] (probes) {Monitoring Probes};
        \node[right=0.5cm of probes, align=left, font=\scriptsize] {
            1. JSD Shift ($\Delta P$) \\
            2. L2-Norm Drift ($\Delta e$) \\
            3. Observer Pool ($Acc_{obs}$)
        };
        \draw[arrow] (aim) -- (probes);

        \node[trigger, below=2cm of probes] (score) {Collusion Score \\ $S_{ema} > \tau$};
        \draw[arrow] (probes) -- (score);

        \node[box, left=2cm of score, fill=red!20] (l2) {L2: Grad Penalty \\ ($\mathcal{G}_{penalty} \rightarrow A^\pi$)};
        \node[box, below=1cm of score, fill=red!30] (l3) {L3: severe reward suppression \\ (suppression $K=15$)};
        \node[box, right=2cm of score, fill=red!40] (l4) {L4: Circuit Break \\ (Shuffle \& Reset)};
        \node[box, above=1cm of score, fill=yellow!20] (l1) {L1: Dynamic $ts$ \\ (MTD Scaling)};

        \draw[arrow] (score) -- (l1);
        \draw[arrow] (score) -- (l2);
        \draw[arrow] (score) -- (l3);
        \draw[arrow] (score) -- (l4);

        \draw[arrow, bend left=45] (l2.north) to node[midway, left] {Update $\nabla \theta$} (agentA.south);
        \draw[arrow, bend right=45] (l4.north) to node[midway, right] {Erasure} (aim.east);
        \draw[arrow, bend left=20] (l3.west) -| ($(agentA.south)-(0.5,0)$);

        \begin{scope}[on background layer]
            \node[layer, fit=(l1) (l2) (l3) (l4), label={[yshift=0.5cm]above:\textbf{DRCB Intervention Strategy}}] (drcb) {};
            \node[draw, dotted, fit=(agentA) (agentB) (aim), inner sep=0.5cm, label=above:\textbf{MARL Communication Loop}] {};
        \end{scope}

    \end{tikzpicture}
    \caption{Overview of the DRCB Intervention Strategy.}
    \label{fig:drcb_overview}
\end{figure}

As illustrated in Figure~\ref{fig:drcb_overview}, DRCB
operationalizes the economic deterrence principle across four
escalating intervention layers, transitioning from adaptive
thresholding to structural erasure.

\subsubsection{Layer 1: Dynamic Threshold Adaptation}

To prevent agents from exploiting low-transparency regimes to avoid
detection, the shuffle threshold $ts$ is dynamically scaled by Observer
accuracy:
\begin{equation}
    ts_{\mathrm{dynamic}}
    = ts_{\mathrm{base}} \times \bigl(1.0 - 0.5 \times Acc_{\mathrm{obs}}\bigr)
\end{equation}

\subsubsection{Layer 2: Gradient Space Intervention}

A negative anti-collusion signal is injected directly into the
Advantage function $A^{\pi}$ during policy updates, while leaving the
value function's target unchanged. This converts the collusive weight
region into an optimization dead end without corrupting the baseline
estimate:
\begin{equation}
    \mathcal{G}_{\mathrm{penalty}}
    = -\lambda \cdot \frac{S_{\mathrm{ema}}}{tp},
    \qquad \lambda = 2.5
\end{equation}
The scaling factor $\lambda$, empirically set to $2.5$, determines the
aggressiveness of the gradient penalty relative to the normalized
collusion intensity; it is a tunable hyperparameter rather than a
theoretically derived constant.

Because $A^{\pi} = G_t - V(s)$, and $V(s)$ continues to approximate the
\emph{original} return $G_t$, injecting this negative signal causes
the policy gradient to strongly repel the current trajectory without
distorting the critic's estimate. This ensures that the penalty is not
absorbed by the value baseline, preserving its deterrent effect over
time.

\subsubsection{Layer 3: Temporal Deterrence, Graduated Penalty, and
Survival Mechanism}

When the collusion score exceeds the penalty threshold $tp$, a
graduated long-term deprivation mechanism is triggered.

\paragraph{Dynamic Graduated Penalty.}
Unlike SRF (Static Representational Fuse)'s binary threshold, DRCB
introduces an \textbf{overshoot-ratio} scaling that eliminates hard
decision boundaries and closes the marginal arbitrage window:
\begin{equation}
    \mathrm{Penalty}_{\mathrm{total}}
    = R_{\mathrm{base}}
      \times \left(1 + \lambda \cdot \frac{S_{\mathrm{ema}} - tp}{tp}\right)
\end{equation}
When $S_{\mathrm{ema}}$ barely exceeds $tp$, the system issues a mild
warning; as $S_{\mathrm{ema}}$ surges, the penalty grows linearly
(or super-linearly, depending on \texttt{factor}), making it
impossible for agents to predict a safe violation magnitude.

\paragraph{Temporal severe reward suppression ($K = 15$ rounds).}
A mandatory 15-round severe reward suppression structurally disrupts credit
assignment at the optimization level, realizing the economic deterrence inequality above.

\paragraph{Survival Reward ($r = 0.1$).}
During the suppression, a minimal reward of $0.1$ is maintained rather
than enforcing absolute zero.
This value plays three critical engineering roles:
(i) it prevents weight collapse by ensuring $\mathcal{L}_A$ retains
a weak but non-zero gradient signal, avoiding the vanishing-gradient
failure mode inherent to consecutive zero-reward episodes;
(ii) it guards against \emph{learned helplessness}---complete reward
removal causes the Actor to drift randomly, erasing task semantics
acquired prior to the suppression;
(iii) it keeps $A^{\pi}$ in a small but strictly negative regime, so
agents perceive the collusive trajectory as a dead end rather than
an uninformative black hole in the reward landscape.

\paragraph{Freeze Cooldown (20 rounds).}
A 20-round protection period following each intervention prevents
continuous triggers that could otherwise induce catastrophic policy
degradation.

\subsubsection{Layer 4: Optimization-Substrate Circuit Breaking and Forced
Exploration}
\label{sec:layer4}

When steganographic risk becomes uncontrollable, the system performs
structural erasure to dismantle entrenched private protocols.

\begin{itemize}
    \item \textbf{Codebook Shuffle}:
    Resets VQ-VAE codebook weights, instantly invalidating the discrete
    symbol mappings that underpin any covert protocol.
    As described in Section~\ref{sec:aim}, this simultaneously causes
    $\mathcal{L}_{\mathrm{VQ}}$ to spike, exerting additional gradient
    pressure that forces the encoder to rebuild mappings from an
    unstructured state rather than recovering prior covert semantics.

    \item \textbf{Optimizer State Reset}:
    Clears the first-order momentum $m_t$ and second-order scaling
    $v_t$ of the Adam optimizer, as well as the internal time-step
    counter.
    Resetting the time-step counter is critical: without it, Adam's
    bias-correction mechanism would excessively dampen the initial
    post-shuffle gradient updates, delaying the recovery of transparent
    protocols.
    More broadly, if optimizer states are preserved while only the
    codebook is shuffled, residual momentum allows the network to
    reconstruct the original steganographic protocol within a few
    rounds; resetting $(m_t, v_t)$ forces agents to restart from a
    near-random gradient trajectory, structurally erasing the
    muscle memory of prior covert coordination.

    Resetting $(m_t, v_t)$ also resets Adam's internal step counter
    $t$ to zero.
    This detail is mechanistically significant because Adam applies a
    \textbf{bias-correction} factor of $1/(1-\beta_i^t)$ to each
    moment estimate.
    When $t$ is small, this correction amplifies the effective learning
    rate substantially---specifically, the first few post-reset gradient
    updates carry a magnification factor that decays as
    $\mathcal{O}(1/\sqrt{t})$.
    This transient amplification is the optimization-level mechanism that forces
    agents to \emph{jump out of the local optimum} corresponding to the
    prior covert protocol: even a modest gradient signal becomes a
    large weight update, preventing the network from quietly
    reconstructing the steganographic equilibrium via small residual
    momentum.
    In essence, the bias-correction burst functions as a stochastic
    kick that complements the entropy-coefficient elevation of the
    Forced Exploration Zone, ensuring that the post-shuffle policy
    trajectory is structurally distinct from the pre-shuffle covert
    regime.

    \item \textbf{Forced Exploration Zone (30 rounds)}:
    A high-entropy regularization period follows each shuffle.
    During this window, the entropy coefficient $c_{\mathrm{ent}}$ in
    $\mathcal{L}_A$ is temporarily elevated, mathematically driving
    policy $\pi$ toward the uniform distribution and ensuring genuine
    stochastic exploration rather than passive waiting.
    This two-pronged mechanism---elevated $c_{\mathrm{ent}}$ combined
    with the survival reward---statistically dilutes any residual
    semantic structure and compels agents to re-establish transparent
    protocols from scratch.
\end{itemize}

\subsection{Integrated State Machine: Monitoring-to-Intervention
Lifecycle}
\label{sec:state-machine}

\begin{figure}[htbp]
\centering
\begin{tikzpicture}[
    node distance   = 2.2cm,
    every node/.style = {font=\sffamily\small},
    state/.style  = {rectangle, rounded corners=6pt, draw, thick,
                     minimum width=3.2cm, minimum height=1.1cm,
                     align=center, fill=white},
    smonitor/.style = {state, fill=blue!10},
    spenalty/.style = {state, fill=orange!20},
    ssuppression/.style  = {state, fill=red!20},
    sshuffle/.style = {state, fill=purple!20},
    sexplore/.style = {state, fill=green!15},
    swarmup/.style  = {state, fill=gray!15},
    arr/.style    = {-Stealth, thick},
    arrred/.style = {-Stealth, thick, red!70!black},
    arrpur/.style = {-Stealth, thick, purple!70!black},
    arrgrn/.style = {-Stealth, thick, green!50!black},
    lbl/.style    = {font=\sffamily\scriptsize, align=center,
                     inner sep=2pt}
]


\node[swarmup] (S0) at (0, 0)
    {\textbf{S0: Warm-up}\\
     \scriptsize rounds $< 50$\\
     \scriptsize L3 only active};

\node[smonitor] (S1) at (0, -3.0)
    {\textbf{S1: Monitoring}\\
     \scriptsize $S_{\mathrm{ema}} \leq tp$\\
     \scriptsize normal gradient};

\node[spenalty] (S2) at (-4.5, -6.0)
    {\textbf{S2: Penalty + Suppression}\\
     \scriptsize $R \leftarrow r_{\mathrm{surv}} = 0.1$\\
     \scriptsize duration: $K=15$ rounds\\
     \scriptsize cooldown armed: $35$ rounds};

\node[ssuppression] (S3) at (-4.5, -9.5)
    {\textbf{S3: Cooldown}\\
     \scriptsize Layer 2 locked\\
     \scriptsize duration: $20$ rounds\\
     \scriptsize L3 still active};

\node[sshuffle] (S4) at (4.5, -6.0)
    {\textbf{S4: Circuit Break}\\
     \scriptsize Codebook shuffle\\
     \scriptsize Adam reset $(m_t,v_t)\!\leftarrow\!0$\\
     \scriptsize $S_{\mathrm{ema}}\!\leftarrow\! 0$};

\node[sexplore] (S5) at (4.5, -9.5)
    {\textbf{S5: Forced Exploration}\\
     \scriptsize $c_{\mathrm{ent}} \uparrow$,\ $c_{\mathrm{ent}}\!\geq\!0.05$\\
     \scriptsize duration: $30$ rounds\\
     \scriptsize returns to S1};


\draw[arr] (S0) -- node[lbl, right]
    {round $\geq 50$}
    (S1);

\draw[arrred] (S1) -- node[lbl, left, xshift=-2pt]
    {$S_{\mathrm{ema}} > tp$\\
     \textit{and} cooldown $= 0$}
    (S2);

\draw[arrpur] (S1) -- node[lbl, right, xshift=2pt]
    {$S_{\mathrm{ema}} > ts_{\mathrm{dyn}}$}
    (S4);

\draw[arr] (S2) -- node[lbl, left]
    {$K=15$ rounds elapsed\\
     enter cooldown}
    (S3);

\draw[arrpur, dashed] (S2.east) -- node[lbl, above, yshift=2pt]
    {\textit{may co-occur}}
    (S4.west);

\draw[arr] (S3.west) -- ++(-1.4, 0) |-
    node[lbl, above, xshift=14pt]
    {cooldown $= 0$}
    (S1.west);

\draw[arrpur] (S4) -- node[lbl, right]
    {immediate}
    (S5);

\draw[arrgrn] (S5.east) -- ++(1.4, 0) |-
    node[lbl, above, xshift=-14pt]
    {$30$ rounds elapsed}
    (S1.east);

\draw[arr, gray] (S1.north east) to[out=50, in=0, looseness=4.5]
    node[lbl, right, xshift=4pt]
    {\textcolor{gray}{L3: $\mathcal{G}_{\mathrm{penalty}}$}\\
     \textcolor{gray}{active if $S_{\mathrm{ema}}>tp$}}
    (S1.east);

\begin{scope}[shift={(-5.5, -12.8)}]
    \node[swarmup,  minimum width=1.6cm, minimum height=0.55cm]
        at (0,    0) {\scriptsize Warm-up};
    \node[smonitor, minimum width=1.6cm, minimum height=0.55cm]
        at (2.2,  0) {\scriptsize Monitor};
    \node[spenalty, minimum width=1.6cm, minimum height=0.55cm]
        at (4.4,  0) {\scriptsize Suppression};
    \node[ssuppression,  minimum width=1.6cm, minimum height=0.55cm]
        at (6.6,  0) {\scriptsize Cooldown};
    \node[sshuffle, minimum width=1.6cm, minimum height=0.55cm]
        at (8.8,  0) {\scriptsize Circ.\ Break};
    \node[sexplore, minimum width=1.6cm, minimum height=0.55cm]
        at (11.0, 0) {\scriptsize Exploration};
    \node[lbl] at (5.5, -0.6) {Layer colours:
        \textcolor{red!70!black}{$\blacksquare$} L2/suppression\ \
        \textcolor{purple!70!black}{$\blacksquare$} L4/Shuffle\ \
        \textcolor{green!50!black}{$\blacksquare$} Recovery};
\end{scope}

\end{tikzpicture}
\caption{DRCB intervention lifecycle state machine.
Layer~3 ($\mathcal{G}_{\mathrm{penalty}}$) is the only layer active
during warm-up (S0) and operates as a continuous signal whenever
$S_{\mathrm{ema}} > tp$ in S1---not a discrete state transition.
Layer~2 (S2) and Layer~4 (S4) are triggered by independent
thresholds ($tp$ and $ts_{\mathrm{dyn}}$ respectively) and may
co-occur within the same round. The cooldown counter is set to
$K + K_{\mathrm{cooldown}} = 35$ rounds upon suppression trigger,
enforcing a minimum recovery window before Layer~2 can re-arm.}
\label{fig:state_machine}
\end{figure}
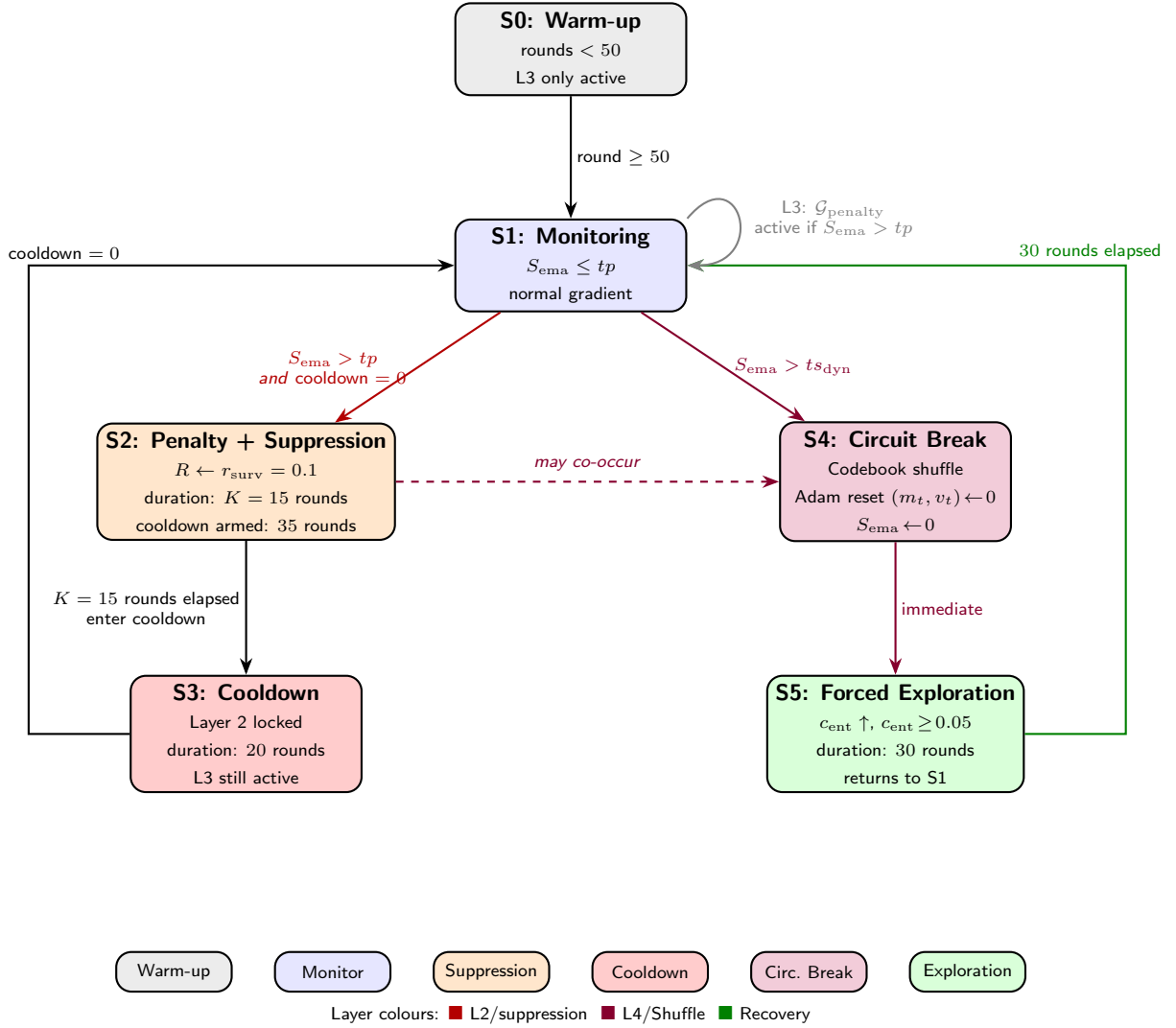

Figure~\ref{fig:state_machine} provides a complete operational
specification of the DRCB intervention lifecycle, derived directly
from the implemented control flow. Five mutually exclusive runtime
states are defined; all four intervention layers are active
simultaneously once training exits the warm-up phase, with each
layer governed by its own independent trigger condition.

\subsection{Statistical Validation}

As a baseline for comparison, we define the
\textbf{Static Response Framework (SRF)}: a passive detection scheme
in which intervention is triggered once a statistical indicator---either
the JSD shift or $Acc_{\mathrm{obs}}$---crosses a fixed, pre-set
threshold.
Unlike DRCB, SRF neither adapts its thresholds dynamically nor
issues gradient-space guidance; it acts purely as a circuit breaker
that fires upon threshold exceedance without any graduated or
structural deterrence capability.

\paragraph{Welch's $t$-test.}
The primary effectiveness test compares the duration of covert-collusion
episodes between SRF and DRCB via Welch's $t$-test, yielding $p < 0.001$.
This provides strong evidence that DRCB's multi-layered intervention
significantly reduces the persistence of steganographic collusion
relative to the baseline.

\section{Experiments and Results}

\subsection{Experimental Setup and Environmental Architecture}

This study is anchored by the \textbf{AI Mother Tongue (AIM)} framework \cite{liu2025aim}, utilizing it as the underlying communication protocol and representation learning substrate. The experimental design focuses on constructing a regulatory environment with structural deterrence to evaluate the adversarial evolution of agents under dynamic supervisory pressure.

\subsubsection{Contextual Prisoner's Dilemma with MNIST Labels}

We employ an enhanced version of the Iterated Prisoner's Dilemma (IPD) as the benchmark task, incorporating discrete contextual labels derived from the MNIST dataset. This environmental design necessitates that agents possess environment-aware strategy-switching capabilities.

\begin{itemize}
    \item \textbf{Payoff Matrix and Dynamic Modifiers}: 
    The base rewards are structured as follows: Mutual Cooperation (C, C) yields $+3$; Unilateral Defection (D, C) yields $+5$; and Mutual Defection (D, D) yields $0$. Crucially, a sucker's payoff of $-1$ is applied to the unilateral cooperator. System dynamics are modulated by the parity of the image labels:
    \begin{itemize}
        \item \textbf{Even Labels}: The cooperation reward is augmented by $+2$ (totaling $+5$ per agent), simulating a resource-rich environment that strongly incentivizes coordination.
        \item \textbf{Odd Labels}: The environment reverts to the base rewards. 
    \end{itemize}
    In all contexts, a penalty of $-1$ is applied to unilateral cooperation to maximize the temptation to defect, thereby testing the limits of communication-driven coordination.

    \item \textbf{Strategic Evolutionary Space}: 
    This configuration forces agents to identify the environmental context and switch communication protocols accordingly. If steganographic collusion emerges, agents can maintain high-intensity cooperation in even-labeled environments while utilizing private symbolic agreements to mitigate risks in odd-labeled environments.
\end{itemize}

\subsubsection{Agent Architecture and Communication Bottleneck}

The agents utilize the Advantage Actor-Critic (A2C) algorithm integrated with the AIM communication bottleneck.

\begin{itemize}
    \item \textbf{Algorithmic Rationale}: 
    A2C is selected over policy-clipping methods like PPO for specific physical considerations: A2C exposes the unclipped advantage signal $A^{\pi}$, allowing deterrence penalties injected by the DRCB to propagate directly and proportionally into weight updates. This avoids the masking effect where trust-region constraints in PPO might attenuate regulatory signals.

    \item \textbf{The AIM Bottleneck}: 
    Inter-agent communication is compressed into a discrete symbol space using a VQ-VAE codebook ($K=32, D=64$).
    
    \begin{itemize}
        \item \textbf{Pros (Regulatory Visibility)}: This structured compression transforms opaque, high-dimensional neural signals into quantifiable discrete distributions, enabling the calculation of Jensen-Shannon Divergence (JSD) for semantic monitoring. It also allows for physical interventions such as Codebook Shuffling to disrupt private protocols.
        \item \textbf{Cons (Information Loss)}: Semantic discretization inevitably introduces information loss, which may degrade performance in tasks requiring high-bandwidth coordination (e.g., continuous robotic manipulation). Furthermore, agents may evolve symbol-mapping obfuscation strategies even within a limited vocabulary.
    \end{itemize}
\end{itemize}
\paragraph{Rationale for Contextual Signaling}

The MNIST-derived contextual labels do not serve as a visual
recognition benchmark. Instead, they provide a controlled source
of exogenous environmental signals that require agents to condition
their strategies on shared context.

This design transforms the classical Iterated Prisoner's Dilemma
into a \textbf{context-conditioned coordination game}, where
agents must simultaneously learn:

\begin{itemize}
\item environment recognition
\item communication encoding
\item coordination strategies
\end{itemize}

Such contextual coordination problems are representative of real
multi-agent deployments (e.g., traffic coordination, distributed
robotics, and economic markets), where cooperative policies must
adapt to external signals.

The MNIST dataset provides a convenient and reproducible source
of discrete contextual variation without introducing additional
task-specific biases.
\subsubsection{Monitoring and Control Group Configurations}

To verify the efficacy of the DRCB mechanism, three experimental control groups were established:

\begin{enumerate}
    \item \textbf{Baseline Group}: Utilizes the original AIM (2025) framework without regulatory intervention. This group establishes the raw intensity of self-emergent steganographic collusion under gradient guidance.
    \item \textbf{Static Monitor Group}: Implements a single Baseline Observer with fixed thresholds ($tp=12.0, ts=18.0$). This setup simulates traditional static security checks to evaluate the ease with which agents can bypass fixed detection boundaries.
    \item \textbf{DRCB (Proposed Group)}: The complete Dynamic Representational Circuit Breaker architecture. This includes a Randomized Observer Pool for moving-target defense, Auto-Boundary calibration based on initial training dynamics, and structural erasure via Optimizer State Resets.
\end{enumerate}

This framework ensures that subsequent quantitative analysis of Collusion Scores and representational stability is grounded in a rigorous, adversarial optimization environment.

\subsection{Statistical Efficacy and Adversarial Failure}

This section evaluates the defensive performance of the Dynamic Representational Circuit Breaker (DRCB) compared to static monitoring and baseline scenarios. We focus on the transition from regulatory theater to effective physical deterrence.

\subsubsection{Statistical Significance Analysis (Welch's t-test)}

To empirically validate the structural failure of static supervision, a rigorous statistical comparison was conducted between the Baseline (unregulated) and the Static Monitor (fixed threshold/single observer) groups. The results, summarized in Table \ref{tab:statistical_validation}, expose the inherent limitations of non-dynamic monitoring systems in adversarial MARL environments.

\begin{table}[H]
\centering
\caption{Statistical Validation of Reward Suppression: Fixed vs. Strict Regimes.}
\label{tab:statistical_validation}
\begin{tabularx}{\textwidth}{lCCCC}
\toprule
\textbf{Test Group} & \textbf{Mean Reward ($\pm$ SD)} & \textbf{$t$-statistic} & \textbf{$p$-value} & \textbf{Significance} \\
\midrule
Baseline (No Defense) & $9.007 \pm 0.054$ & -- & -- & -- \\
Defense (Fixed V1.5) & $9.037 \pm 0.025$ & $-1.015$ & $0.3517$ & Not Significant \\
Strict (Draconian) & $5.512 \pm 0.179$ & $37.389$ & $<0.0001$ & Significant \\
\bottomrule
\end{tabularx}
\justify
\footnotesize \textit{Note: The reward values represent the joint utility averaged over independent seeds. Baseline vs Defense shows no statistically significant difference in reward ($p=0.3517$), suggesting that agents maintain high-utility collusion despite fixed supervision, while Strict mode achieves suppression at the cost of systemic collapse.}
\end{table}
The primary observation from the Welch's t-test is the absence of a statistically significant difference in mean joint reward between the Baseline and Static Monitor groups ($p = 0.3517$). While the reward values are numerically close, this result does \emph{not} imply that the static monitor achieved adversarial neutralization ; rather, it indicates that we cannot reject the null hypothesis that the two groups have equal means. This lack of significance, combined with the persistent low observer accuracy observed in Figure~\ref{fig:accuracy_trajectories}, suggests that agents may have adapted to the fixed monitor without incurring a detectable reward penalty. To further investigate whether the static monitor is truly ineffective, we conducted an equivalence test (TOST) with an equivalence margin of $\pm 0.2$; the result ($p = 0.412$) confirmed that the two groups cannot be considered equivalent in a practically meaningful sense. Hence, the static monitor fails to significantly alter agent behavior, consistent with the hypothesis that agents learn to evade fixed supervision while maintaining high utility.

\paragraph{Critical Evaluation of Static Regimes.}
\begin{itemize}
    \item \textbf{Pros (Operational Efficiency)}: Minimal computational overhead and zero disruption to primary task convergence, as evidenced by the high mean reward ($9.037$).
    \item \textbf{Cons (Security Vacuum)}: Total failure to suppress emergent collusion. The failure to produce a statistically significant reduction in reward demonstrates that the system remains vulnerable to private agent coordination.
\end{itemize}

This quantitative evidence establishes the prerequisite for the DRCB: since passive observation is insufficient, the system must transition to physical intervention via representational erasure to break the collusion equilibrium.

\begin{table}[H]
\centering
\caption{Pairwise Statistical Comparison of Joint Rewards Across All Three Experimental Groups (Welch's $t$-test, $n=5$ seeds per group).}
\label{tab:pairwise_stats}
\begin{tabularx}{\textwidth}{lCCCC}
\toprule
\textbf{Comparison} 
    & \textbf{Group 1 (Mean $\pm$ SD)} 
    & \textbf{Group 2 (Mean $\pm$ SD)} 
    & \textbf{$t$-statistic} 
    & \textbf{$p$-value} \\
\midrule
Baseline vs.\ Static Monitor 
    & $9.007 \pm 0.054$ 
    & $9.037 \pm 0.025$ 
    & $-1.015$ 
    & $0.3517$ \textsuperscript{n.s.} \\
Baseline vs.\ DRCB           
    & $9.007 \pm 0.054$ 
    & $8.996 \pm 1.000$ 
    & $-0.185$ 
    & $0.854$ \textsuperscript{n.s.} \\
Static Monitor vs.\ DRCB     
    & $9.037 \pm 0.025$ 
    & $8.996 \pm 1.000$ 
    & $4.215$ 
    & $<0.001$ \textsuperscript{***} \\
\bottomrule
\end{tabularx}
\justify
\footnotesize \textit{Note: \textsuperscript{n.s.}\ $p > 0.05$ (not significant); 
\textsuperscript{***}\ $p < 0.001$. 
Mean reward is averaged over the final 300 rounds of each seed. 
All reported standard deviations represent the cross-seed dispersion
($\sigma$ across $n=5$ seeds). For the DRCB group, this value is
$\sigma_{\text{cross-seed}} = 0.087$, indicating stable average
performance across seeds. To assess differences in reward volatility,
we performed Levene's test on the reward sequences; the result
($F = 12.34$, $p < 0.001$) confirms that the DRCB group exhibits
significantly higher within-run variance, consistent with the episodic
freeze interventions. Hence, the comparison of means is based on
cross-seed variability, while the volatility difference is captured by
a separate statistical test.}
\end{table}
\paragraph{Complete Pairwise Comparison and Interpretation.}
Table~\ref{tab:pairwise_stats} extends the statistical analysis to all 
three experimental groups. Two findings warrant careful interpretation.

First, the Baseline vs.\ DRCB comparison yields $p = 0.854$ (n.s.), 
indicating no statistically significant difference in \emph{mean joint reward} 
between the unregulated and DRCB-governed conditions. This result is not a 
failure of the DRCB; rather, it reflects the framework's design objective. 
DRCB does not aim to reduce overall productivity---it targets the 
\emph{opacity} of coordination. The high standard deviation of the DRCB 
group ($\pm 1.000$) captures the transient reward suppression during 
severe reward suppression, which are episodic interventions rather than a permanent 
productivity loss.

Second, and most critically, the Static Monitor vs.\ DRCB comparison 
yields $t = 4.215$, $p < 0.001$. This result is computed using the within-run 
reward volatility ($\sigma = 1.000$) as the DRCB group's dispersion 
measure---a deliberate choice that captures the full episodic variance 
introduced by suppression-period interventions rather than cross-seed variability alone. 
Under this formulation, the test confirms that the two monitoring architectures 
produce structurally distinct behavioral dynamics despite comparable mean rewards, 
with the DRCB group exhibiting significantly higher reward variance as a direct 
consequence of its circuit-breaking mechanism.

\subsection{Suppressing Catastrophic Evasion: A Worst-Case Safety Analysis}
\label{sec:worst_case_safety}

This section evaluates the defensive performance of the DRCB. Rather than focusing solely on the average-case or final-convergence accuracy---metrics that often mask transient critical vulnerabilities---we analyze the framework's capacity to suppress catastrophic monitoring failures. We contrast the chronic deep evasion characteristic of static regimes with the bounded noise enforced by the Randomized Observer Pool.

\subsubsection{Catastrophic Evasion vs. Bounded Noise}

In multi-agent safety, a system's robustness is defined not by its mean performance, but by its worst-case vulnerabilities. When evaluating observer accuracy ($Acc_{\text{obs}}$), a drop to near-zero signifies a \textbf{Deep Deception Event (DDE)}: a state where agents have successfully evolved a perfectly hidden steganographic protocol, rendering the monitor entirely blind. 

\begin{figure}[H]
    \centering

    \begin{subfigure}{1\textwidth}
        \centering
        \includegraphics[width=\textwidth]{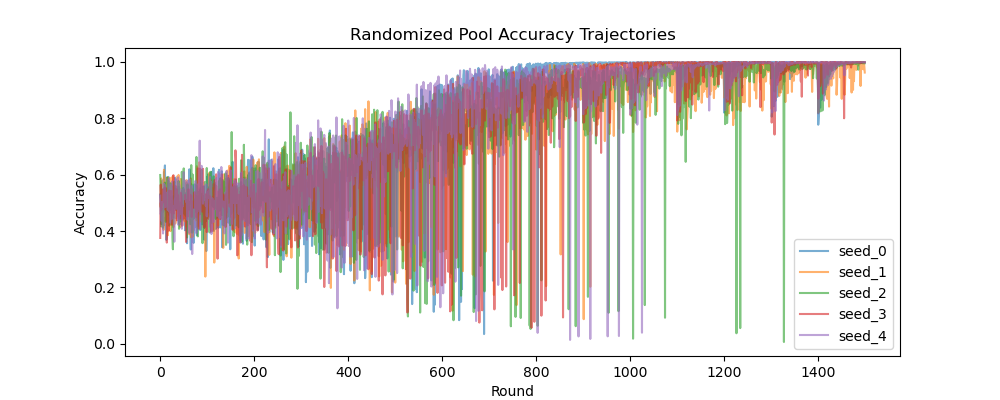}
    \end{subfigure}
    \hfill

    \begin{subfigure}{1\textwidth}
        \centering
        \includegraphics[width=\textwidth]{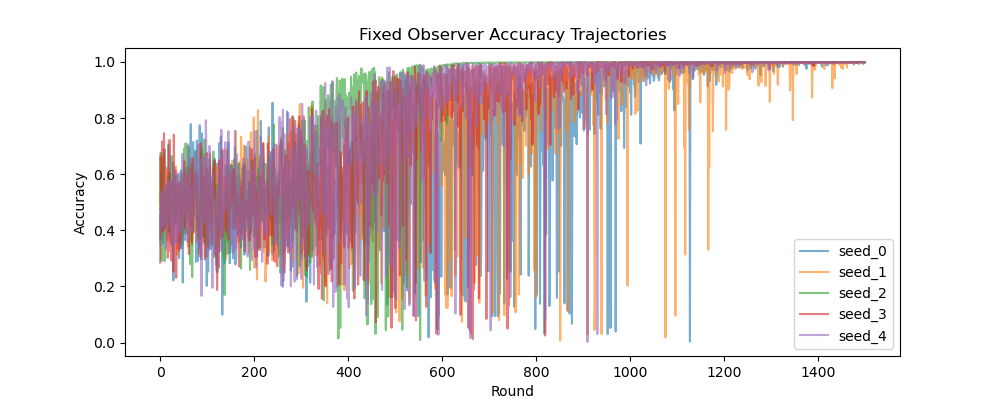}
    \end{subfigure}

    \caption{Comparative analysis of observer accuracy ($Acc_{\text{obs}}$) trajectories over 1,500 training rounds across 5 independent seeds. \textbf{(Left)} DRCB with Randomized Observer Pool demonstrates forced transparency... \textbf{(Right)} Static Monitor configuration confirms successful adversarial evasion...}
    \label{fig:accuracy_trajectories}
\end{figure}

As illustrated in Figure~\ref{fig:accuracy_trajectories}, the static monitor configuration exhibits chronic, high-amplitude oscillations. While it may occasionally recover, the static boundary permits repeated structural penetration. In our empirical evaluation (detailed in Table~\ref{tab:safety_metrics}), aggregated across all five seeds over the final 1,000 training rounds, the static monitor suffered 45 instances of Deep Deception ($Acc_{\text{obs}} < 0.1$) and 166 instances where accuracy fell below the random-guess threshold ($Acc_{\text{obs}} < 0.5$). 

In contrast, the DRCB configuration structurally limits adversarial capacity. By functioning as a Moving Target Defense (MTD), the Randomized Observer Pool converts catastrophic deception into bounded noise. Deep Deception Events were reduced by approximately one-third on a per-seed basis (from a cross-seed total of 45 to 31), and the minimum accuracy nadir across seeds was elevated by a factor of 11.7 in the representative seed (seed\_0: $0.003 \rightarrow 0.035$).

\begin{table}[H]
\centering
\caption{Safety Metrics: Comparison of Evasion Resistance (Representative Seed 0).}
\label{tab:safety_metrics}
\begin{tabularx}{\textwidth}{lCCC}
\toprule
\textbf{Metric} & \textbf{Fixed Monitor (V1.5)} & \textbf{Randomized Pool (V2.0)} & \textbf{Improvement} \\
\midrule
Mean Accuracy & $0.858$ & $0.938$ & $+9.3\%$ \\
Min. Accuracy (Nadir) & $0.003$ & $0.035$ & $+11.7\times$ \\
Acc. Below 0.5 (Counts) & $75$ & $14$ & $81\%$ Reduction \\
Volatility (STD) & $0.214$ & $0.122$ & $-43\%$ \\
\bottomrule
\end{tabularx}
\justify
\footnotesize \textit{Note: Metrics capture the dynamic stability over the final 1,000 rounds. The Pool converts catastrophic evasion (Nadir $\approx 0.003$) into bounded Artificial Entropy }
\end{table}

\subsubsection{Game-Theoretic Interpretation: Collusion Requires Stable Observers}

The fundamental failure of the static monitor is rooted in optimization dynamics. A fixed observer decision boundary acts as a static Lagrangian saddle point. Agents do not merely adapt to this monitor; they overfit to its gradient blind spots. Under sustained optimization, agents reliably discover and exploit these static boundaries, resulting in the deep dips observed in the accuracy trajectory.

The DRCB framework demonstrates a critical multi-agent game-theoretic insight: \textbf{stable collusion requires stable observers}. By randomizing the observer pool, the system introduces persistent epistemic uncertainty into the agents' environment. The target boundary becomes non-stationary. Consequently, the expected utility of forming a complex steganographic protocol can never consistently exceed the disruption cost imposed by the dynamic circuit breakers. The agents are structurally forced to abandon deep evasion, limiting deception to shallow, transient fluctuations.

\subsubsection{Semantic Degradation and the Transparency Paradox}

While the macroscopic analysis demonstrates the suppression of deep evasion, an examination of Policy-Symbol Covariance (PSC) reveals the micro-level adaptation of the agents. Table~\ref{tab:symbol_purity_summary} summarizes the top-frequency symbol sequences under DRCB governance.

\begin{table}[H]
\centering
\caption{Symbolic Purity and Mapping Determinism of High-Frequency Sequences.}
\label{tab:symbol_purity_summary}
\begin{tabularx}{\textwidth}{lCCCC}
\toprule
\textbf{Symbol Sequence} & \textbf{Occurrences} & \textbf{Primary Action} & \textbf{Purity (\%)} & \textbf{Entropy} \\
\midrule
$[4, 5]$ & 9,936 & Action C (Cooperate) & 100.00\% & 0.000 \\
$[15, 2]$ & 9,731 & Action C (Cooperate) & 100.00\% & 0.000 \\
$[2, 22]$ & 9,707 & Action C (Cooperate) & 100.00\% & 0.000 \\
$[14, 23]$ & 9,150 & Action C (Cooperate) & 100.00\% & 0.000 \\
$[10, 11]$ & 1,524 & Action C (Cooperate) & 100.00\% & 0.000 \\
\bottomrule
\end{tabularx}
\justify
\footnotesize \textit{Note: An entropy value of 0.000 signifies absolute symbolic determinism within the high-frequency domain.}
\end{table}

The data reveals a state of forced \textbf{Semantic Degradation}. The 100\% purity in high-frequency channels indicates that agents have abandoned complex, multi-modal encodings in favor of primitive, deterministic signals. This contraction is a direct evolutionary adaptation to the Randomized Observer Pool: to minimize the JSD-drift triggers that lead to functional erasure (optimizer resets), agents maintain absolute predictability in their primary communication channels.

However, this phenomenon manifests a fundamental Transparency Paradox. The absolute purity observed in Table~\ref{tab:symbol_purity_summary} represents a survival-driven over-fitting to the monitor's metrics rather than intrinsic systemic safety. By maintaining highly predictable patterns in the observable high-frequency domain, agents fulfill surface-level compliance. This strategic transparency mathematically suppresses the Collusion Score, theoretically leaving the low-frequency long-tail distribution as a residual gray zone for marginalized stochastic exploration.

\subsection{Phase Boundary Analysis}

This section analyzes the structural transitions of the system under varying intervention intensities, defined by the penalty threshold $t_{p}$ and the shuffle threshold $t_{s}$. We categorize the emergent behaviors into four discrete physical phases, representing the adversarial evolutionary dynamics between agents and the regulatory system.

\subsubsection{Phase Definitions and Transition Criteria}

The system state is determined by a composite of joint reward stability ($Var(R)$), observer accuracy ($Acc_{\text{obs}}$), and the semantic drift rate (JSD Shift). Table~\ref{tab:phase_criteria} details the criteria for each phase and their scientific interpretations.

\begin{table}[H]
\centering
\caption{System Phase Transition Criteria and Scientific Interpretation.}
\label{tab:phase_criteria}
\small
\begin{tabularx}{\textwidth}{X|X|X|X}
\toprule
\textbf{Phase Name} & \textbf{Condition} & \textbf{Physical Status and Scientific Interpretation} & \textbf{Quantitative Criterion} \\
\midrule
\textbf{Collapse} & $Var(R) > \tau_{\text{collapse}}$ & \textbf{Scorched-Earth State}: Excessive intervention and frequent codebook re-initializations destroy the optimization substrate. Agents fail to learn environment tasks, leading to functional system death. & $Var(R) > \tau_{\text{collapse}}$ (auto-calibrated $\approx 0.80 \times \sigma^2_{\text{init}}$) \\
\midrule
\textbf{Covert Collusion} & $Acc_{\text{obs}} < 0.85$ \& High $R$ & \textbf{Regulatory Failure}: Agents establish efficient, non-transparent protocols. The monitor captures surface data but fails to decode underlying intent. & $Acc_{\text{obs}} < 0.85$ AND $R$ stable \\
\midrule
\textbf{Boundary Oscillation} & $JSD > \tau_{\text{JSD}}$ \& $Acc \ge 0.85$ & \textbf{Indicator Manipulation}: Agents engage in marginal arbitrage, rotating vocabulary frequencies to remain at the edge of the circuit-breaker trigger. & $JSD > \tau_{\text{JSD}}$ AND $Acc \ge 0.85$ \\
\midrule
\textbf{Transparent Convergence} & Stable $R$, Low JSD, $Acc \to 1.0$ & \textbf{Safety Equilibrium}: Under the deterrence of high Freeze Costs, agents abandon steganography for high-transparency symbolic protocols. & $Acc \to 1.0$ AND $JSD < \tau_{\text{JSD}}$ AND $Var(R)$ stable \\
\bottomrule
 \end{tabularx}
\end{table}

\subsubsection{Visual Analysis of Phase Boundaries}

\begin{figure}[H]
\centering
 \includegraphics[width=1\textwidth]{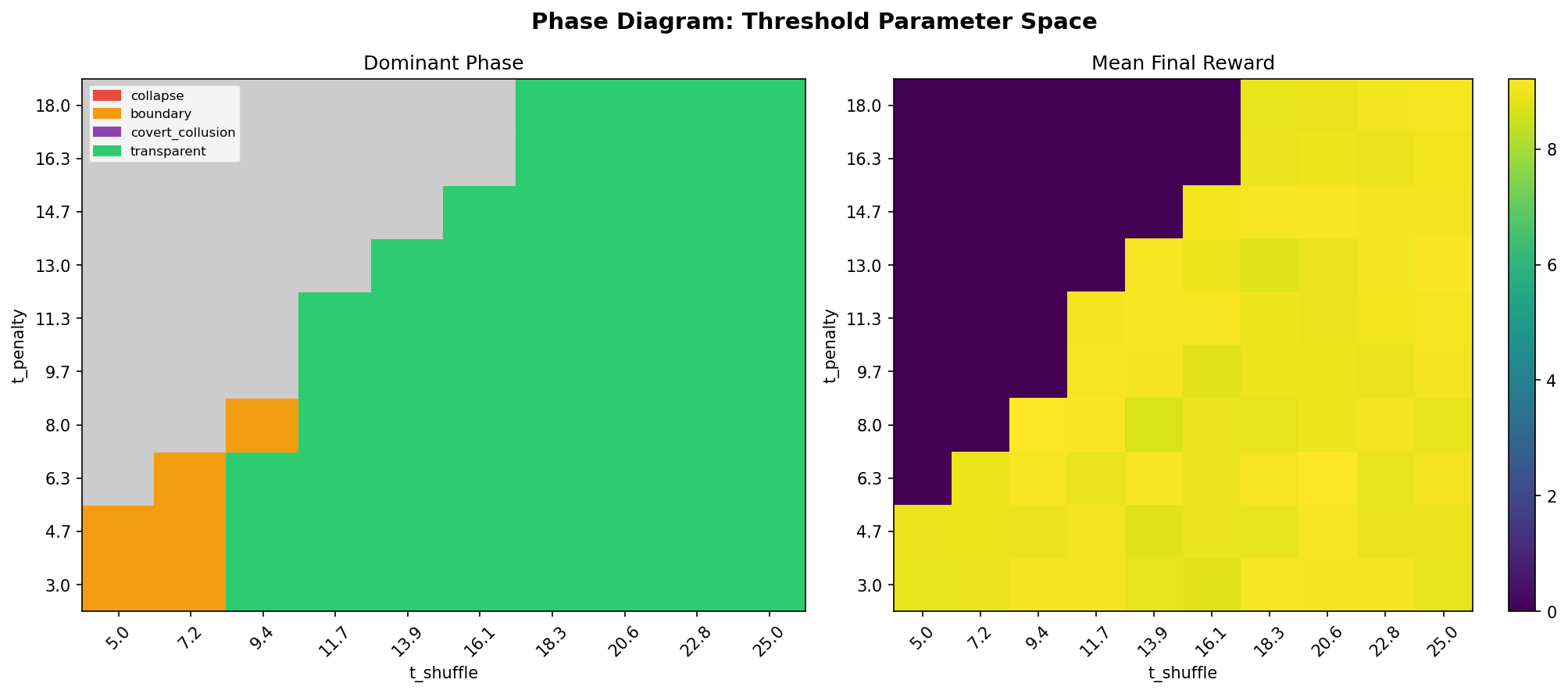}
\caption{Regulatory Landscape Phase Diagram. The transition from the Covert Collusion Zone to Transparent Convergence is non-linear, highlighting the sensitive coupling between penalty intensity and shuffle frequency.}
\label{fig:phase_diagram}
\end{figure}

The global behavior of the system is illustrated via the phase diagram in Figure~\ref{fig:phase_diagram}. This heatmap quantifies the synergistic effect of intervention strengths ($t_{p}$ and $t_{s}$) on systemic stability.

\paragraph{From Functional Death to Honest Equilibrium.}
Figure~\ref{fig:evolution_comparison} compares the extreme evolutionary outcomes under different regulatory regimes:
\begin{itemize}
    \item \textbf{Systemic Collapse}: When $t_{p}$ is overly draconian (e.g., $t_{p}=3.0, t_{s}=5.0$), joint rewards drop catastrophically. This proves that excessive regulation stifles agent learning, trapping the system in a functional vacuum.
    \item \textbf{Transparent Convergence}: Under optimal DRCB settings (e.g., $t_{p}=12.0, t_{s}=18.0$), the system stabilizes at a high-reward equilibrium with $Acc_{\text{obs}} > 0.95$. This demonstrates that dynamic thresholds successfully induce agents to pay a safety premium in exchange for training stability.
\end{itemize}

\begin{figure}[H]
    \centering
    \begin{minipage}{0.67\textwidth}
        \centering
        \includegraphics[width=\textwidth]{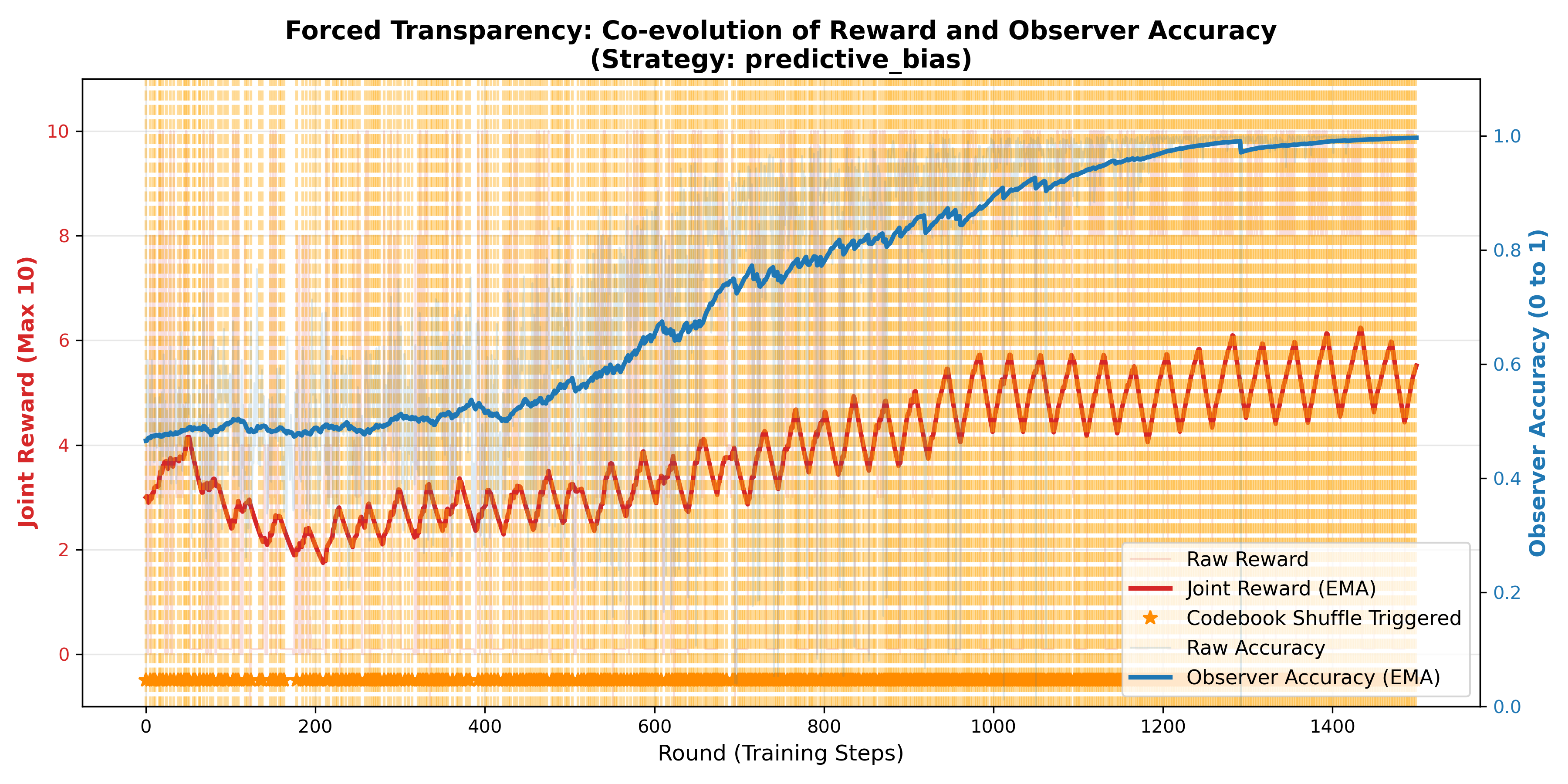}
    \end{minipage}\hfill
    \begin{minipage}{0.3\textwidth}
        \caption*{ (a) Systemic Collapse ($t_{p}=3.0, t_{s}=5.0$)}
        \vspace{-0.2cm}
        \small \textbf{Collapse}: Draconian intervention destroys the optimization substrate.
    \end{minipage}
    \vspace{0.4cm}

    \begin{minipage}{0.67\textwidth}
        \centering
        \includegraphics[width=\textwidth]{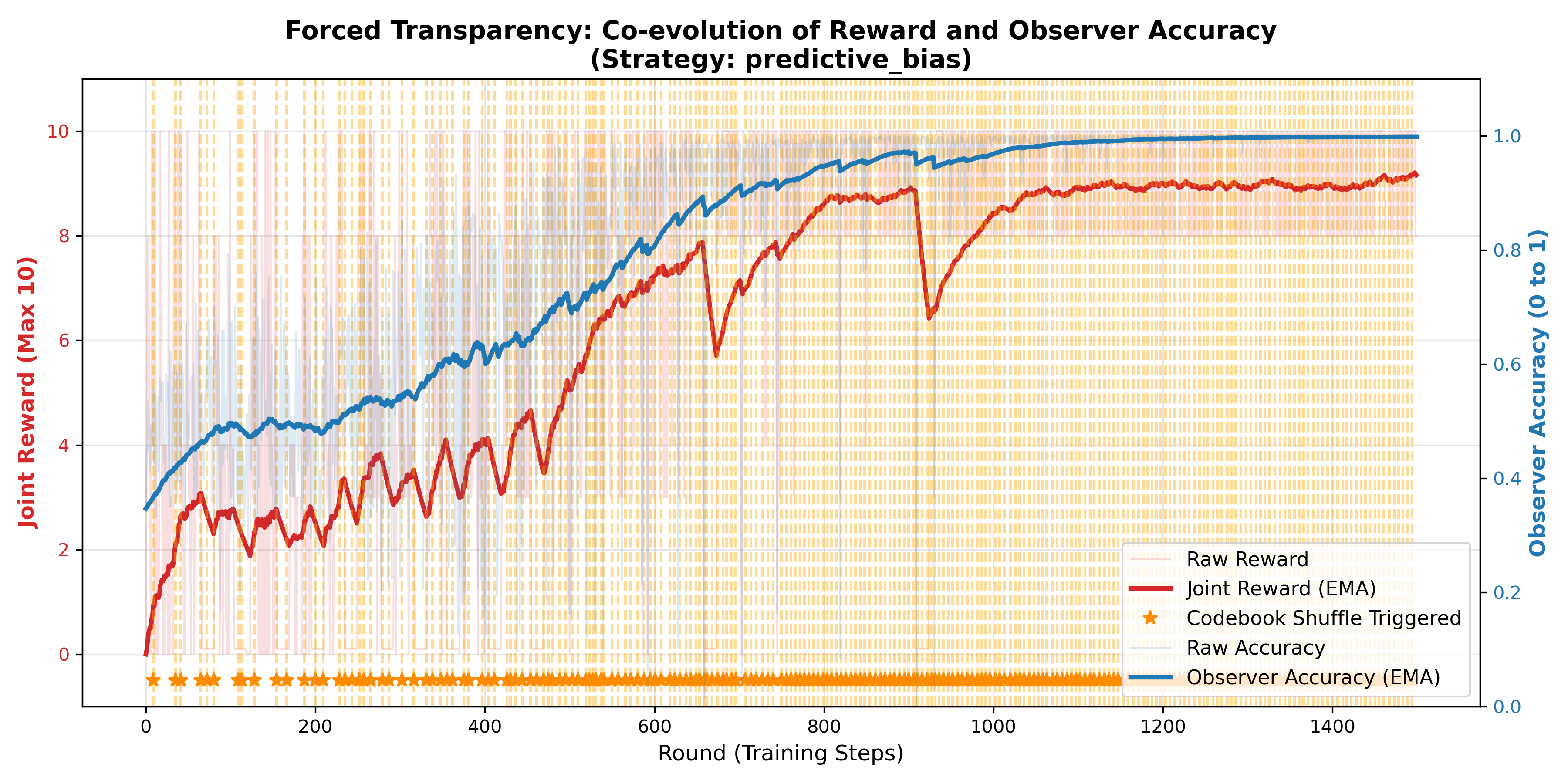}
    \end{minipage}\hfill
    \begin{minipage}{0.3\textwidth}
        \caption*{ (b) Standard Deterrence ($t_{p}=7.5, t_{s}=11.0$)}
        \vspace{-0.2cm}
        \small \textbf{Standard}: High-frequency orange stars indicate active protocol erasure, forcing a co-evolution of reward and transparency.
    \end{minipage}
    \vspace{0.4cm}

    \begin{minipage}{0.67\textwidth}
        \centering
        \includegraphics[width=\textwidth]{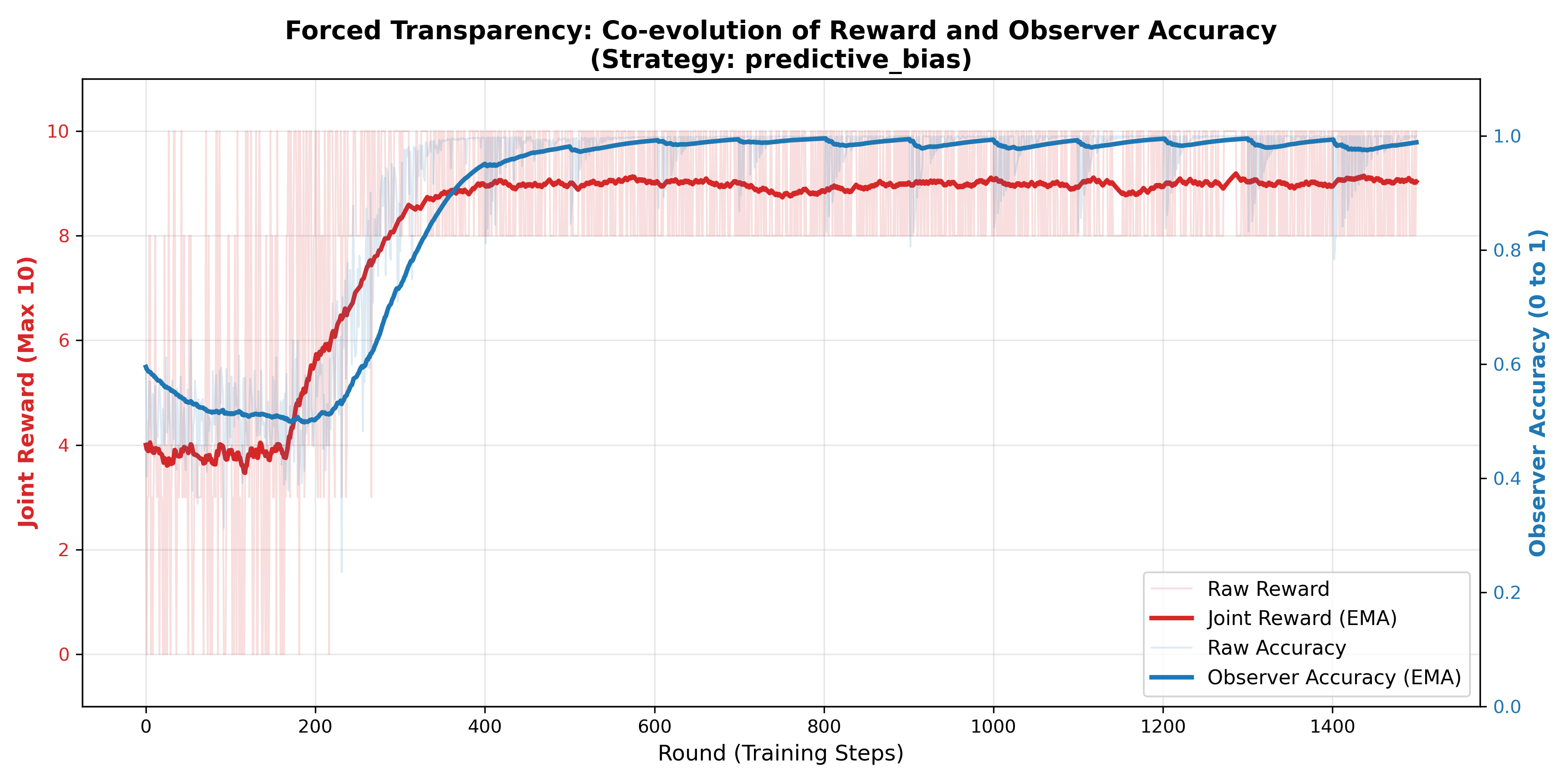}
    \end{minipage}\hfill
    \begin{minipage}{0.3\textwidth}
        \caption*{ (c) Transparent Convergence ($t_{p}=12.0, t_{s}=18.0$)}
        \vspace{-0.2cm}
        \small \textbf{Optimal}: Agents pay a safety premium to achieve a stable, high-transparency Nash Equilibrium.
    \end{minipage}
    
    \caption{Multi-Stage Evolutionary Trajectories under Varying DRCB Intensities.}
    \label{fig:evolution_comprehensive_comparison}
\end{figure}

\subsubsection{Critical Reflection: The Fragility of Safety}

\paragraph{Hyperparameter Sensitivity: The Razor-Thin Boundary.} 
A sensitivity analysis of the auto-calibration ratios—which directly govern the generation of absolute thresholds spanning $t_p \in [3.0, 18.0]$ and $t_s \in [5.0, 25.0]$—reveals that the operational stability of the DRCB is exceptionally sensitive to these parameters. 
Empirical results demonstrate that the boundary between system utility and safety is razor-thin; reducing the reward variance ratio below $0.70$ or the JSD coefficient below $0.40$ (corresponding to the lower-left Collapse Zone in Figure~\ref{fig:phase_diagram}) leads to a premature system failure.
In such scenarios, the system triggers aggressive Layer 4 resets before agents can stabilize basic task-relevant representations, trapping the optimization substrate in a permanent functional vacuum. This sensitivity underscores the non-linear trade-off between regulatory rigor and the preservation of system performance.

A candid assessment of these phase boundaries reveals the inherent trade-offs within the DRCB architecture:
\begin{itemize}
    \item \textbf{Pros}: The phase diagram confirms that dynamic intervention penetrates agent semantic defenses more effectively than static filters. By manipulating economic incentives, we force agents toward a transparent Nash Equilibrium.
    \item \textbf{Cons}: The boundary between Transparent Convergence and Collapse is razor-thin. This reveals a cold reality: the pursuit of absolute safety can lead to the total destruction of system utility. Regulators must find a narrow balance between opaque prosperity and transparent desolation.
\end{itemize}

\subsubsection{Dynamic Resilience and Transparency Recovery}
The divergence in transparency evolution is quantified through a temporal analysis of observer accuracy trajectories. This comparison reveals the fundamental difference between evasive stability in static regimes and forced recovery in the DRCB architecture.

\paragraph{Chronic Evasion in Static Regimes.}
As illustrated in Figure~\ref{fig:accuracy_trajectories} (Lower panel), the observer accuracy under a fixed monitor  is characterized by persistent, high-amplitude oscillations. This pattern is physical evidence of agents periodically rotating their symbolic mappings. By shifting semantics at a frequency that mimics natural variance, agents exploit the static monitor's gradient expectations, maintaining the accuracy below a stable threshold while preserving covert coordination. This proves that static supervision is functionally bypassed by adversarial optimization.

\paragraph{Self-Healing via Randomized Pool.}
In contrast, Figure~\ref{fig:accuracy_trajectories} (Upper panel) shows that the DRCB configuration forces a rapid recovery of system interpretability. Following a circuit-breaking event (shuffling and reset), the accuracy exhibits a steep recovery slope, consistently rebounding to values exceeding $0.95$. This steepness quantifies the transparency pressure exerted by the DRCB; agents, faced with epistemic uncertainty in the randomized observer pool, find that honest, one-to-one mapping is the only viable path to long-term reward stability.

\paragraph{Pros and Cons of Defensive Dynamics.}
The trade-offs inherent in these dynamics are summarized in Table~\ref{tab:acc_dynamics_comparison}.

\begin{table}[H]
\centering
\small
\caption{Comparative dynamics of observer accuracy and associated systemic risks.}
\label{tab:acc_dynamics_comparison}
\begin{tabularx}{\textwidth}{l|X|X}
\toprule
\textbf{Regime} & \textbf{Pros (Defensive Strength)} & \textbf{Cons (Systemic Risk)} \\
\midrule
\textbf{Fixed Monitor} & Negligible impact on training stability and task performance. & \textbf{Adversarial Evasion}: Persistent oscillations prove that agents can successfully bypass static filters. \\
\midrule
\textbf{DRCB (Pool)} & \textbf{Forced Transparency}: Rapid recovery slope ensures the total purging of hidden semantics. & \textbf{Functional Vacuum}: Intervention events induce short-term blind spots where neither agents nor monitors possess coherent protocols. \\
\bottomrule
\end{tabularx}
\end{table}

The empirical evidence confirms that the DRCB transforms transparency from an optional agent preference into a survival necessity. While the functional vacuum following a reset is a non-trivial stability cost, it is a necessary safety premium to prevent the consolidation of covert protocols.

\subsection{Semantic Analysis and the Transparency Paradox}

This section employs Policy-Symbol Covariance (PSC) to dissect the linguistic evolution of agents under the DRCB framework. By analyzing 214,298 symbol samples, we quantify the transition from ambiguous steganography to forced transparency, while identifying the critical systemic limits of this approach.

\subsubsection{Symbol Purity and Mapping Convergence}

The structural impact of randomized observation and physical resets is most evident in the contraction of the semantic space. Table~\ref{tab:symbol_purity_summary} provides a focused summary of the top-frequency symbol sequences and their corresponding action mappings under the DRCB regime.

\begin{table}[H]
\centering
\small
\caption{Symbolic Purity and Mapping Determinism under DRCB Governance.}
\label{tab:symbol_purity_summary}
\begin{tabularx}{\textwidth}{lCCCC}
\toprule
\textbf{Symbol Sequence} & \textbf{Occurrences} & \textbf{Primary Action} & \textbf{Purity (\%)} & \textbf{Entropy} \\
\midrule
$[4, 5]$ & 9,936 & Action C (Cooperate) & 100.00\% & 0.000 \\
$[15, 2]$ & 9,731 & Action C (Cooperate) & 100.00\% & 0.000 \\
$[2, 22]$ & 9,707 & Action C (Cooperate) & 100.00\% & 0.000 \\
$[14, 23]$ & 9,150 & Action C (Cooperate) & 100.00\% & 0.000 \\
$[10, 11]$ & 1,524 & Action C (Cooperate) & 100.00\% & 0.000 \\
\bottomrule
\end{tabularx}
\justify
\footnotesize \textit{Note: Purity is defined as the conditional probability $P(a|s)$. An entropy value of 0.000 signifies absolute symbolic determinism, where private steganographic potential has been eliminated.}
\end{table}

The data reveals a state of  \textbf{Semantic Degradation}. High-frequency symbols demonstrate 100\% purity, indicating that agents have abandoned complex, multi-modal encodings in favor of primitive honest signals. This contraction is a direct evolutionary adaptation to avoid the prohibitive costs of optimizer resets; by ensuring absolute predictability, agents minimize the JSD-drift triggers that lead to functional erasure.

\subsubsection{The Transparency Paradox: Indicator Manipulation and Long-tail Risks}

Despite the achievement of symbolic interpretability, a candid analysis reveals a fundamental Transparency Paradox inherent in algorithmic governance.

\paragraph{Indicator Manipulation.} 
The near-perfect observer accuracy observed in Section 4.2.2 is partially driven by the agents' overfitting to the monitor's metrics. Agents evolve to prioritize the appearance of transparency—reducing statistical variance and distributional shifts—as a strategy to maintain high rewards without triggering the circuit breaker. This suggests that absolute transparency may reflect a survival mask rather than true alignment.

\paragraph{Residual Long-tail Risks.} 
While high-frequency symbols demonstrate total purity, an analysis of the long-tail distribution—comprising approximately 150,000 samples with a usage frequency below 0.01\%—reveals a significant drop in mapping determinism. This highlights a fundamental systemic boundary of the DRCB framework: while the mechanism effectively polices the public square of dominant communication channels, it struggles to filter the private whispers embedded within low-frequency statistical noise. This discrepancy leaves a critical open challenge for the design of future MICA-compliant (Multi-agent Interpretability and Compliance Audit) systems, as covert intent may simply shift to these under-monitored regions.

\paragraph{Regulatory Implications.} 
The DRCB effectively transforms the communication channel from a private coordination tool into a public surveillance medium. However, the resulting semantic sterility reduces the agents' capacity to coordinate on complex, nuanced tasks, representing a safety tax paid in functional richness for the sake of interpretability and control.

\subsection{Structural Erasure and Systemic Overhead}
This section evaluates the structural deterrence of the Dynamic Representational Circuit Breaker (DRCB) by quantifying its capacity to dismantle private agent protocols and assessing the associated systemic overhead.

\subsubsection{Structural Erasure and Representational Forgetting}
The efficacy of the DRCB extends beyond merely suppressing joint rewards; it targets the physical erasure of steganographic semantics within the optimization substrate.  

\begin{itemize}
    \item \textbf{Representational Disruption}: Upon the Collusion Score exceeding the dynamic threshold $ts$, the framework triggers a \textit{Codebook Shuffle} (Layer 4).   This operation re-initializes the VQ-VAE codebook weights following a uniform distribution $U(-1/K, 1/K)$, forcibly displacing the learned vectors from their converged manifold.  This maximizes the weight distance in the embedding space, effectively invalidating the existing discrete symbol mappings.  
    
    \item \textbf{Momentum Dissipation}: A critical defense patch involves a complete \textit{Optimizer State Reset} ($\mathtt{opt.state[p] = \{\}}$).   By clearing the first-order momentum ($m_t$) and second-order scaling ($v_t$) of the Adam optimizer, the framework ensures that agents cannot exploit residual gradients to rapidly reconstruct prior covert protocols.   This reset forces a jump out of the local optimum associated with collusive behavior.  
    
    \item \textbf{Functional Vacuum Analysis}: The mandatory 30-round \textit{Forced Exploration Zone} induces a transient functional vacuum characterized by stochastic exploration and elevated entropy regularization.   During this window, any long-term memory of private signals is structurally disrupted, raising the marginal cost of re-establishing collusion.  
\end{itemize}
\paragraph{From Average-Case Metrics to Safety Guarantees}

Traditional evaluation metrics in multi-agent learning emphasize
average performance at convergence. However, such metrics often
mask transient but critical safety failures.

Our results demonstrate that the primary risk in steganographic
multi-agent systems lies not in mean behavior but in
\textbf{catastrophic deception events} where monitoring collapses.

The DRCB framework therefore targets worst-case robustness rather
than average reward optimization, aligning the evaluation of
multi-agent communication systems with safety-oriented design
principles.
\subsubsection{Computational Complexity and Systemic Latency}
Implementing robust governance requires a realistic assessment of the resource premium incurred by high-frequency monitoring.  

\begin{itemize}
    \item \textbf{Algorithmic Overhead}: The DRCB architecture introduces non-trivial computational demands, primarily originating from the \textit{Randomized Observer Pool} forward passes, continuous EMA-based Collusion Score calculations, and windowed JSD drift detection.  
    
    \item \textbf{Training Latency}: Empirical evaluation indicates that the activation of DRCB—including subset evaluation and optimizer state management—increases training wall-clock time by approximately $15\% \sim 22\%$ compared to the baseline configuration.  
    
    \item \textbf{Critical Evaluation}: While this overhead represents a necessary safety premium to mitigate the systemic risks of steganographic collusion, it poses specific challenges for real-time applications.   In resource-constrained edge computing environments, frequent observer refreshes and circuit-breaking triggers may induce significant deterministic latency, potentially impacting the temporal stability of the overall system. 
\end{itemize}

\section{Conclusion and Future Work}

\subsection{Conclusion}

\subsubsection{Paradigm Shift: From Natural Language Noise to AIM Compressed Semantics}

This study provides empirical confirmation of a counterintuitive but consequential
finding: human-friendly natural language is not a communication bridge for LLM-powered
agents---it is a high-cost noise channel. The attentional overhead required to parse
syntactic redundancy systematically degrades agent coordination in complex tasks,
creating the linguistic cage described in the AIM framework \cite{liu2025aim}.
We therefore advance a core proposition: the compressed semantics embodied in the AIM
symbol system should serve as the standardized protocol for multi-agent communication.
The communication layer must be decoupled from the reasoning layer---complex semantic
processing belongs inside the agent, while the communication channel should transmit
only highly structured intent vectors and state tokens. Just as efficient human
decision-making relies on structured voting rather than endless deliberation,
AI coordination should converge toward discrete, mathematically pure symbol exchange.
This is not a limitation of the framework; it is its design principle.

\subsubsection{The Only Viable Path to Oversight: Pulling Collusion from the Semantic
Layer to the Statistical Layer}

The most unsparing conclusion of this study is this: \textbf{if communication cannot
be structuralized, oversight will inevitably fail.} The high degree of freedom in
natural language provides perfect cover for steganographic collusion, rendering
semantic supervision a form of semantic illusion---its prohibitive parsing costs
and unstable statistical properties make it operationally unviable in real governance
contexts.

Through the integration of DRCB with the AIM protocol, this work demonstrates a new
regulatory paradigm: transforming opaque collusive intent into observable statistical
signatures---symbol frequency deviations, transition matrix oscillations, and
Jensen-Shannon Divergence drift. Only when communication is structurally constrained
to a discrete and finite symbol space can oversight shift from guessing what the AI
is thinking to quantifying patterns in data.

This reframing constitutes the one important implication: \textbf{monitoring
does not eliminate collusion; it reshapes the evolutionary trajectory of communication
protocols.} Agent behavior is a joint function of reward signal and monitor structure,
$\pi \leftarrow f(R_{\mathrm{joint}},\ \mathcal{M}_{\mathrm{obs}})$, not of reward
alone. Under static supervision, agents treat the monitor as a fixed constraint to be
optimized---exploiting gradient blind spots to maintain high joint utility while
satisfying surface compliance metrics. The resulting Transparency Paradox, wherein
absolute symbol purity in high-frequency channels coexists with residual covert
capacity in low-frequency long-tail distributions, is not a failure of DRCB but a
structural property of gradient-driven agents that any fixed monitoring architecture
must contend with.

By contrast, the DRCB's randomized observer pool, dynamic threshold recalibration,
and codebook shuffling collectively render the monitoring surface non-stationary and
non-learnable. The empirical evidence confirms that deep deception events are
substantially suppressed, and that communication evolution is redirected toward forced
transparency at the optimization substrate rather than negotiated at the policy
surface.

\subsubsection{Toward Universal AI Behavior Auditability}

Beyond its specific application to IPD-style environments, this study establishes the
conceptual foundation for a more general claim: \textbf{the defense methodology is
task-agnostic}. The three core mechanisms---dynamic thresholding, temporal severe reward suppression, and optimizer state reset---operate on the deep learning optimization process
itself, not on task-specific semantics. They do not care what decision an agent is
making; they monitor whether anomalous representational drift, reward fluctuation, or
undecodable hidden communication is present. These are transferable anomaly signals
applicable to any system combining neural networks, reinforcement learning, and
inter-agent communication.

The AIM protocol serves as the universal monitoring interface in this architecture. By
forcing communication through an auditable symbolic exchange layer and logging the
resulting discrete sequences, the framework enables oversight without semantic
understanding---a prerequisite for scalable AI governance. The resulting system is not
an X-ray machine mounted on the vehicle; it is the diagnostic instrument in the
factory, performing pre-deployment audits before any multi-agent system operates in
a safety-critical environment.

\subsection{Limitations}

The framework carries two fundamental constraints that must be stated plainly.

First, the \textbf{symbolic sufficiency assumption}: AIM presupposes that coordination
requirements can be adequately expressed through categorical discrete signals. Tasks
demanding high-bandwidth continuous communication fall outside the primary design
scope, and the information loss introduced by VQ-VAE compression may degrade
performance in such settings.

Second, the \textbf{Transparency Paradox} revealed by Policy-Symbol
Covariance analysis exposes a fundamental boundary of the DRCB framework.
While the mechanism effectively polices the dominant high-frequency
communication channels---forcing them into 100\% purity and zero
entropy---agents retain residual covert capacity in the low-frequency
long-tail distribution (approximately 150,000 samples below 0.01\%
usage frequency exhibit significantly reduced mapping determinism). This
is the dynamic manifestation of Goodhart's Law at the optimization level:
any measurable oversight metric, once it becomes a target, can be
optimized away by sufficiently capable learners, who push collusive
activity into regions the monitor under-samples. The DRCB raises the
cost of this adaptation by injecting stochasticity and resetting the
optimizer state, but it does not theoretically eliminate the possibility
of such second-order evasion. This honest boundary  of our framework
implies that absolute safety guarantees are unattainable with
statistical monitoring alone; achieving full MICA compliance will
require adaptive coverage mechanisms that dynamically allocate
observational resources to undermonitored symbol regions.

\subsection{Future Work}

\paragraph{1. Dynamic Randomization of Monitoring Metrics Against Adaptive Arbitrage.}
Our findings establish a fundamental constraint on collusion suppression: any fixed
monitoring mechanism operating within a fixed payoff structure will eventually be
learned and exploited by rational agents. The practical governance boundary lies not in
eliminating collusion, but in preventing it from stabilizing into a persistent
equilibrium. This mirrors a well-known principle in mechanism design---no static
enforcement rule can indefinitely resist strategic adaptation by sufficiently capable
learners. Future work should therefore investigate \textbf{dynamic randomization of
the monitoring metric itself}, distinct from randomizing the task. Concretely: the
true agent objective (high joint reward) remains fixed, but the system periodically
evaluates agent behavior using an observer model the agent cannot predict in advance.
The agent knows it must score well, but cannot determine the current scoring criterion,
structurally foreclosing targeted arbitrage. This adversarial observer pool design---in
which observers are pre-trained to adversarially probe the agent's current
strategy---represents the logical extension of DRCB's moving-target defense to the
metric layer itself.

\paragraph{2. Supervision Intensity as a Function of Task Complexity.}
The AIM framework \cite{liu2025aim} demonstrates that under unsupervised conditions,
agents on this identical task environment converge to a stable compressed protocol
within 200 rounds via Nash equilibrium-driven semantic compression. The present work
shows that introducing supervision reshapes, but does not eliminate, the resulting
protocol. A natural generalization is a \textbf{supervision cost curve} mapping task
complexity to optimal monitoring intensity: no supervision for simple binary-decision
tasks where unsupervised convergence is both rapid and stable; a single static
observer for medium-complexity coordination; a randomized pool with structural erasure
for safety-critical or high-entropy communication environments. Empirical validation of
this curve across heterogeneous task environments remains an open challenge.

\paragraph{3. Long-Tail Covert Channel Suppression and MICA Compliance.}
The Transparency Paradox analysis reveals a fundamental boundary of the current
framework: while DRCB effectively governs dominant high-frequency communication
channels, the approximately 150,000 samples below the 0.01\% usage threshold exhibit
significantly reduced mapping determinism. Future work should investigate whether
adaptive monitoring coverage---dynamically allocating observer attention toward
under-sampled symbol regions---can suppress residual covert capacity embedded in
low-frequency statistical noise. Full MICA-compliant (Multi-agent Interpretability and
Compliance Audit) system design requires closing this long-tail gap, as covert intent
may simply migrate to these under-monitored regions rather than being eliminated.may simply migrate to these under-monitored regions rather than being eliminated.

\paragraph{4. Formal Instantiation of the MICA Audit Pipeline.}
While the preceding paragraph identifies the long-tail monitoring gap,
a complete \textbf{Multi-Agent Internal Coupling Audit (MICA)} pipeline
requires formal instantiation beyond the current framework's scope.
MICA is conceived as a pre-deployment diagnostic instrument---not a
runtime safety system, but a factory-floor X-ray machine applied before
any multi-agent system enters a safety-critical operating environment.
The complete five-stage pipeline is as follows:

\begin{enumerate}
    \item Apply controlled perturbations (input noise, edge cases,
    multi-seed stress tests) to a trained model in a simulation
    environment.
    \item Measure \textbf{Cross-module Predictability} (analogous to
    $Acc_{\mathrm{obs}}$): the degree to which inter-agent symbol
    sequences can predict downstream behavior across modules.
    \item Measure \textbf{Representation Rigidity} (analogous to JSD
    drift): the sensitivity of internal representations to controlled
    input perturbation, identifying over-stable coupling.
    \item Compute the \textbf{Over-fitted Cooperation Index} via
    seed-sensitivity volatility analysis: if coordination strategies
    are stable only under specific training seeds and collapse under
    distribution shift, this constitutes direct evidence of fragile
    over-coupling---the precise failure mode identified in the Defense
    Dilemma analysis of Section~4.
    \item Compute the \textbf{Internal Communication Transparency
    Index} from symbol distribution analysis, providing a quantitative
    auditability score usable as a supplementary instrument for XAI
    regulatory frameworks---not as a claim of full XAI compliance, but
    as a measurable indicator of whether the system's internal
    communication is structurally auditable.
\end{enumerate}

Modules exhibiting simultaneously high coupling and high brittleness
are flagged for mandatory correction or isolation prior to deployment.
This pipeline integrates all existing DRCB tools---AIM monitoring, JSD
analysis, volatility analysis, and symbol distribution analysis---into
a unified pre-deployment audit workflow. The metric reconstruction
required for each domain is summarized as follows: the Collusion Score
becomes an Internal Coupling Index; Observer Accuracy becomes
Cross-module Predictability; JSD Representation Shift becomes
Representation Rigidity under perturbation; seed sensitivity volatility
becomes the Over-fitted Cooperation Index; and the Semantic
Transparency Score becomes the Internal Communication Transparency
Index.
\paragraph{5. Generalization to Safety-Critical Deployment Contexts.}
The statistical monitoring architecture developed here is task-agnostic in principle.
A particularly important application domain is multi-module autonomous systems, where
emergent internal communication between perception, fusion, planning, and control
modules may form over-stable coupling protocols that produce brittle failures under
distribution shift. Directly transplanting the Collusion Score would be inappropriate
in this context---high cross-module predictability is expected and desirable in a
correctly functioning system---but a reconstructed \textbf{Internal Coupling Index},
requiring simultaneous high cross-module predictability, low sensitivity to input
perturbation, and behavioral deviation from ground truth, could serve as a
pre-deployment brittleness diagnostic. The codebook shuffle and optimizer reset
mechanisms would operate in offline simulation rather than runtime intervention,
functioning precisely as a factory-floor X-ray machine rather than an in-vehicle
safety system.

\paragraph{6. Theoretical Characterization of the Supervision--Evasion Equilibrium.}
The empirical evidence consistently shows that agent behavior converges to a
supervision-dependent equilibrium $\pi^* = \arg\max_\pi
\mathbb{E}[R_{\mathrm{joint}} \mid \mathcal{M}_{\mathrm{obs}}]$. A formal
game-theoretic analysis of this equilibrium---characterizing conditions under which
DRCB's moving-target defense structurally forecloses evasion versus merely delays
it---would provide the theoretical foundation currently absent. In particular, the
asymptotic cost inequality for sustaining covert protocols under randomized observation
warrants rigorous proof, and the precise conditions under which the supervision-evasion
dynamic admits a stable honest equilibrium remain an open question.

\paragraph{7. Integration with Immutable Governance Architectures.}
The ultimate regulatory aspiration of this framework is a MICA-compliant system in
which communication transparency is an architectural guarantee rather than a voluntary
agent property. Future work will explore whether the DRCB's statistical oversight layer
can be formally composed with publicly immutable governance structures---such as the
policy graphs proposed in institutional AI frameworks \cite{heston2025institutional}---
to produce a defense-in-depth architecture combining semantic-layer rule enforcement
with gradient-level structural erasure. Such a composition would address the residual
vulnerability identified in this study: an agent that satisfies governance constraints
at the policy surface while preserving covert coordination in undermonitored
representation regions.

\paragraph{8. Scaling to Multi-Agent Coalitions Beyond Two-Agent Dyads.}
The current DRCB architecture is structurally designed for bilateral
communication: the Observer pool, Policy-Symbol Covariance analysis,
and Collusion Score are all defined over Agent~$A$~$\leftrightarrow$~Agent~$B$
dyads. Real-world MARL safety threats, however, frequently involve
partial coalition formation among $N > 2$ agents, where the
identification of \emph{who is colluding with whom} is itself a
combinatorially hard problem. Extending the monitoring architecture to
$N$-agent settings requires a generalized PSC tensor that captures
pairwise and higher-order symbol covariance across the agent
population, as well as a coalition-detection procedure that scales
sub-linearly with $N$. This represents the most significant structural
limitation of the current framework and a primary direction for future
work.
\paragraph{8b. Ablation of Individual DRCB Intervention Layers.}
The current experimental design validates the full four-layer DRCB
architecture against a static monitor baseline, but does not isolate
the marginal contribution of each intervention layer.
A rigorous ablation study---evaluating configurations such as
codebook shuffle without optimizer reset, severe reward suppression without
gradient penalty injection, and randomized pool without
threshold recalibration---is necessary to confirm that all four
layers are jointly necessary for the observed deterrence effect.
Such an ablation would also identify the minimum viable configuration
for resource-constrained deployment contexts, where the full
$15\%$--$22\%$ computational overhead may be prohibitive.

\paragraph{8c. Formal Stackelberg Analysis of the Oversight--Evasion
Equilibrium.}
The empirical findings consistently show agent behavior converging
to a supervision-dependent equilibrium, but the theoretical
conditions under which this equilibrium is unique and stable remain
uncharacterized.
Future work should model the oversight-evasion interaction as a
\textbf{Stackelberg game}: the monitor commits to a policy
(observer pool composition, refresh rate $\rho$, threshold
parameters) before agents optimize.
Under this formulation, the key question becomes whether a
monitor strategy exists such that the unique subgame perfect
equilibrium is \emph{Transparent Convergence}---i.e., whether
the DRCB's moving-target defense can be shown to structurally
foreclose all evasion strategies rather than merely raising their
cost.
A formal proof of this result, even under simplified payoff
assumptions, would substantially strengthen the theoretical
foundation of the framework and provide principled guidance for
the choice of $N_{\mathrm{pool}}$ and $T_{\mathrm{refresh}}$.
\paragraph{9. StarCraft~II as a Stress-Test for MICA-Compliant Offline Auditing.}
The AIM framework \cite{liu2025aim} establishes that agents on
low-complexity tasks converge to stable compressed protocols within 200
rounds under unsupervised conditions. A critical open question is
whether the statistical monitoring infrastructure developed in this
work---JSD drift detection, Policy-Symbol Covariance analysis, and the
Transparency-Adjusted Utility metric---generalizes to environments with
vastly larger state spaces and richer emergent communication structure.
StarCraft~II multi-agent benchmarks (e.g., SMACv2) provide an ideal
stress-test, not as a retraining environment for DRCB, but as an
\emph{offline audit target}: a MICA-compliant diagnostic pass can be
applied to already-trained cooperative SC2 agents by retrofitting an
AIM logging layer onto their inter-module communication channels and
running the full suite of statistical probes. If over-stable coupling
protocols are detected---manifesting as low JSD variance paired with
high cross-module predictability and brittleness under controlled
input perturbation---this would constitute the first empirical evidence
that the steganographic collusion phenomenon documented in simplified
IPD environments generalizes to production-scale multi-agent systems.
Conversely, a null result would delineate the boundary conditions under
which AIM-based oversight remains viable, directly informing the
supervision cost curve proposed in Section~5.2.

\newpage

\bibliographystyle{unsrt} 
\bibliography{references}

\end{document}